\documentclass[journal]{IEEEtran}
\usepackage{epsf, psfrag, amssymb, amsfonts, amsmath, cite, enumerate}
\usepackage{graphicx, subfigure, color, bbm, bm}
\usepackage{algorithm}
\usepackage{algorithmic}
\usepackage{graphicx}
\usepackage{wrapfig}
\usepackage{booktabs}       
\usepackage{hyperref}   
\usepackage{mathtools}
\usepackage{enumitem}
\usepackage{multirow, makecell}

\newtheorem{theorem}{Theorem}
\newtheorem{example}{Example}

\newtheorem{definition}{Definition}

\def\psfancypar#1#2{\begingroup\def\par{\endgraf\endgroup\lineskiplimit=0pt}
               \setbox2=\hbox{\large\sc #2}
               \newdimen\tmpht \tmpht \ht2 \advance\tmpht by \baselineskip
               \font\hhuge=Times-Bold at \tmpht
               \setbox1=\hbox{{\hhuge #1}}
               \count7=\tmpht \count8=\ht1
               \divide\count8 by 1000 \divide\count7 by \count8
               \tmpht=.001\tmpht\multiply\tmpht by \count7
               \font\hhuge=Times-Bold at \tmpht
               \setbox1=\hbox{{\hhuge #1}}
               \noindent
                \hangindent1.05\wd1
               \hangafter=-2 {\hskip-\hangindent
               \lower1\ht1\hbox{\raise1.0\ht2\copy1}%
                \kern-0\wd1}\copy2\lineskiplimit=-1000pt}

\newcommand{\beq}{\begin{equation}}
\newcommand{\eeq}{\end{equation}}
\newcommand{\bqa}{\begin{eqnarray}}
\newcommand{\eqa}{\end{eqnarray}}
\newcommand{\bqn}{\begin{eqnarray*}}
\newcommand{\eqn}{\end{eqnarray*}}

\newcommand{\be}{\begin{enumerate}}
\newcommand{\ee}{\end{enumerate}}
\newcommand{\bi}{\begin{itemize}}
\newcommand{\ei}{\end{itemize}}
\newcommand{\bd}{\begin{description}}
\newcommand{\ed}{\end{description}}
\newcommand{\ba}{\begin{array}}
\newcommand{\ea}{\end{array}}
\newcommand{\bde}{\begin{definition}}
\newcommand{\ede}{\end{definition}}
\newcommand{\bex}{\begin{example}}
\newcommand{\eex}{\end{example}}

\def\boxit#1{\vbox{\hrule\hbox{\vrule\kern3pt
        \vbox{\kern3pt#1\kern3pt}\kern3pt\vrule}\hrule}}

\def\reals{ { {\rm  I \kern-0.15em R }  } }
\def\complex{ {\,{{\rm C} \kern-0.50em \raise0.20ex {  |}}\, }}

\def\0bf{{\bf 0}}
\def\1bf{{\bf 1}}
\def\2bf{{\bf 2}}
\def\3bf{{\bf 3}}
\def\4bf{{\bf 4}}
\def\5bf{{\bf 5}}
\def\6bf{{\bf 6}}
\def\7bf{{\bf 7}}
\def\8bf{{\bf 8}}
\def\9bf{{\bf 9}}

\def\Rbf{{\bf R}}

\def\Rxx{\Rbf_{\ssstyle X\kern-.1em X}}

\let\ssstyle=\scriptscriptstyle

\def\Kout{\setbox1=\hbox{\Huge\bf K}\hbox to
1.05\wd1{\hspace{.05\wd1}
}}
\def\Sout{\setbox1=\hbox{\Huge\bf S}\hbox to 1.05\wd1{\hspace{.05\wd1}
}}

\newtheorem{remark}{Remark}

\begin{document}
	\title{On Low Rank Directed Acyclic Graphs and \\ Causal Structure Learning}
	\author{Zhuangyan Fang, Shengyu Zhu, Jiji Zhang, Yue Liu, Zhitang Chen, and Yangbo He
    \thanks{The work of Y.~Liu was partially supported by the National Natural Science Foundation of China (2201629), the work of J.~Zhang was supported in part by the Research Grants Council of Hong Kong under the General Research Fund (13602818), and the work of Y.~He was supported in part by the National Key R\&D Program of China (2022ZD0160303) and the National Natural Science Foundation of China (11971040). F.~Fang and S.~Zhu contributed equally to this work. \emph{Corresponding author: Yue Liu}.}             
    \thanks{F.~Fang was with the School of Mathematical Sciences, Peking University, Beijing, China (e-mail: fangzy\_math@pku.edu.cn).}
 		\thanks{S.~Zhu was with Huawei Noah's Ark Lab, Hong Kong (e-mail: zhushyu@outlook.com).}
            \thanks{J.~Zhang is with the Chinese University of Hong Kong, Hong Kong (e-mail: 	jijizhang@cuhk.edu.hk).}
             \thanks{Y.~Liu is with the Center for Applied Statistics and School of Statistics, Renmin University of China, Beijing, China (e-mail: liuyue\_stats@ruc.edu.cn)}
            \thanks{Z.~Chen is with Huawei Noah's Ark Lab, Hong Kong (e-mail: chenzhitang2@huawei.com).}
            \thanks{Y.~He with  with the School of
Mathematical Sciences, Peking University, Beijing, China (e-mail: heyb@pku.edu.cn).}
 }
	\maketitle
	\begin{abstract}
	Despite several advances in recent years, learning causal structures represented by directed acyclic graphs (DAGs) remains a challenging task in high dimensional settings when the graphs to be learned are not sparse.  In this paper, we propose to exploit a low rank assumption regarding the (weighted) adjacency matrix of a DAG causal model to help address this problem. We utilize existing low rank techniques to adapt  causal structure learning methods to take advantage of this assumption and establish several useful results relating interpretable graphical conditions to the low rank assumption. Specifically, we show that the maximum rank is highly related to hubs, suggesting that scale-free networks, which are frequently encountered in practice, tend to be low rank. Our experiments demonstrate the utility of the low rank adaptations for a variety of data models, especially with relatively large and dense graphs. Moreover, with a validation procedure, the adaptations maintain a superior or comparable performance even when graphs are not restricted to be low rank.  
	
	\end{abstract}

\section{Introduction}\label{sec:intro}
An important goal in many empirical sciences is to discover the underlying causal structures in various domains,  for the purpose of explaining and understanding phenomena, and predicting effects of interventions~\cite{Pearl2009causality}. Due to the relative abundance of passively observed data as opposed to experimental data,  learning causal structures from purely observational data has been vigorously investigated \cite{Peters2017book,spirtes2000causation}. In this context, causal structures are usually represented by directed acyclic graphs (DAGs) over a set of random variables, and existing structure learning methods fall roughly into two classes: constraint-based~\cite{meek1995casual, spirtes2000causation, zhang2008completeness} and score-based~\cite{Chickering1996learning, Chickering2002optimal, Ramsey2017million}. 

Constraint-based methods, such as PC and fast causal inference \cite{spirtes2000causation}, use conditional independence tests to find a skeleton and then determine the edge directions using some orientation rules~\cite{meek1995casual,zhang2008completeness}.   Due to the combinatorial nature of the acyclicity constraint \cite{Chickering1996learning},  traditional score-based  methods mostly rely on
 local heuristics to search for a DAG according to a predefined score function.
Recently, a smooth acyclicity constraint with respect to (w.r.t.)~graph adjacency matrix was introduced in \cite{zheng2018dags}, and the score-based approach on linear data models was then formulated as a continuous optimization problem with least-squares loss. This change of perspective allows using deep learning techniques to model causal mechanisms and has already given rise to several new algorithms for causal structure learning with non-linear data, e.g., \cite{Yu19DAGGNN,Ng2019GAE, Ng2019masked, Lachapelle2019grandag, Zheng2019learning,ke2019learning}, among others. We refer to these algorithms as gradient-based, as the formulated problems are solved using first-order numerical methods. While gradient-based methods represent the current state of the art in many settings, their performance generally degrades when the target DAG is large and relatively dense, as seen from the empirical results  in the above references.

This issue is of course  a challenge to many other approaches. Authors of \cite{Ramsey2017million} proposed fast greedy equivalence search (fast GES) for impressively large problems, but it works reasonably well only when the structure is very sparse. The max-min hill-climbing (MMHC) method \cite{Tsamardinos2006max} relies on local learning methods that often do not perform well when target nodes have large neighborhoods. A potential reason is that dense graphs tend to have high in-degrees that affect the performance of many causal structure learning methods: the search space of PC \cite{spirtes2000causation} and GES grows exponentially in the maximum in-degree  and the sample complexity of the methods proposed in \cite{Ghoshal2017learning, Ghoshal2018learning} also increases fast with the maximum in-degree. How to improve the performance on relatively large and dense  DAGs is therefore an important question. 

In this work, we study the potential of exploiting a kind of \emph{low rank} assumption on the DAG structure to mitigate this problem. The rank of a graph that concerns us is the algebraic rank of its associated weighted adjacency matrix. Similar to the role of a sparsity assumption, the low rank assumption is methodological and is not restricted to a particular DAG learning method. However, unlike sparsity, the graph rank is an algebraic concept---it is much less apparent when DAGs tend to be low rank and how low rank DAGs behave. Thus, besides demonstrating the utility of exploiting a low rank assumption in causal structure learning, a primary goal of this paper is to improve our understanding of the low rank assumption by relating the rank of a graph to its graphical structure. Such results may also help characterize the rank of a graph in certain cases, which in turn inform the choice of rank related hyperparameters for learning algorithms.

{\bf Contributions}\quad To improve our understanding of low rank DAGs, we  establish some lower bounds on the rank of a DAG in terms of simple graphical conditions, which imply necessary conditions for DAGs to be low rank. We also prove that the maximum possible rank of weighted adjacency matrices associated with a directed graph is closely related to  hubs in the graph, which suggests that scale-free networks tend to be low rank. From this result, we derive several graphical conditions to bound the rank of a DAG from above, providing  sufficient conditions for low rank. We then show how to utilize existing low rank techniques to adapt a class of recently developed gradient-based causal structure learning methods with little extra effort. Empirically,  the low rank adaptations are indeed useful, especially with relatively large and dense graphs that tend to have  high in-degrees.  Not only do they outperform the original algorithms when the low rank condition is satisfied, the performance remains very competitive even when graphs are not restricted to be low rank.

{\bf Other Related Work} \quad A low rank assumption is frequently utilized to solve large scale problems in practice, for example, in graph-based applications~\cite{5707106, Smith2012continuous, Zhou2013Hawkes, Liu2013ggm, Meng2014Learning, Yao2016greedy, Benjamin2019robust}, matrix completion and factorization~\cite{wu2010reconstruction, Benjamin2011jmlr, koltchinskii2011, Ling2012, Jain2013, Sainath2013, Cao_2015_ICCV, Davenport2016lowrank}, network sciences~\cite{Hsieh2012lowrank, Huang2013social, Zhang2017link}, etc. In the context of DAG structure learning, the authors of \cite{Barik2019learning,Tichavsky2018representations} focused on \emph{discrete} Bayesian networks and placed the low rank assumption on the parameter space  encoded by conditional probability tables, which is not easily applied to continuous cases. Instead, in this work, we mostly consider continuous variables and place the low rank assumption on the graph structure, which, to the best of our knowledge, has not been studied in the  literature.

The recently developed gradient-based method brings in an opportunity to employ  existing techniques such as matrix factorization (e.g.,~\cite{Koren2009Matrix, Nathan03ICML, Chi2019Nonconvex}) and nuclear norm (e.g.,~\cite{Martin10icml, Gu_2014_CVPR, Fazel2001minimization, Hu2013Fast, gu2017weighted}) for low rank DAG learning, as we will show in Section~\ref{sec:learning}. However, there is a limited understanding of low rank assumption on the DAG structure, as algebraic knowledge about graph rank is  less apparent than structural knowledge. In graph theory, many works have tried to relate graph rank to graph structure. The minimum rank problem studies the minimum rank among all real matrices whose zero–nonzero pattern of \emph{off-diagonal entries} is described by a simple graph~\cite{fallat2007minimum, HOGBEN20101961,MITCHELL2010430}. Here the diagonal entries of real matrices can pick any value (including zero). This problem has not been fully solved, but some bounds characterized by structural information do exist, e.g., in \cite{HOGBEN20101961}. An analogous maximum rank problem, as mentioned by~\cite{fallat2007minimum}, is trivial: the maximum rank among all real matrices with the same zero–nonzero pattern of off-diagonal entries always  equals the number of vertices, which can be achieved by setting a sufficiently large value to the diagonal elements. We comment again that  the diagonal entries of the considered matrices can take any real values in both the minimum and maximum rank problems. In contrast, for DAG learning, we must place constraints on the diagonal entries and fix them at zero. Related to this new setting is \cite{edmonds1967systems} that studied the maximum rank for matrices with a common zero-nonzero pattern (including diagonal entries). In Section~\ref{sec:rank}, this result will be used to relate the maximum possible rank to a more interpretable graphical condition, which further implies several structural properties of low rank DAGs.

%
%
%
%
%
%

{\bf Paper Organization} \quad After introducing preliminaries in Section~\ref{sec:pre}, we present the bounds on graph ranks with interpretable graphical conditions in Section~\ref{sec:rank}. Section~\ref{sec:learning} discusses the adaptations of existing methods to learn low rank DAGs, followed by experimental investigations in  Section~\ref{sec:exp}. Section~\ref{sec:conclusion} concludes the paper. 
	
\section{Preliminaries}\label{sec:pre}
	
We first introduce graph terminology, notations, and recently developed gradient-based causal structure learning methods.
	
	\subsection{Graph Terminology}\label{sec:sec:graph}
	A graph $\mathcal{G}=(\mathbf{V}, \mathbf{E})$ consists of a vertex set $\mathbf{V}=\{X_1, X_2,\cdots,X_d\}$ and an edge set $\mathbf{E}\subset\mathbf{V}\times\mathbf{V}$. We particularly focus on directed (acyclic) graphs. We use $\mathrm{pa}(\mathbf{S}, {\cal G})$, $\mathrm{ch}(\mathbf{S}, {\cal G})$, and $\mathrm{adj}(\mathbf{S}, {\cal G})$ to denote the union of all parents, children, and adjacent vertices of the nodes in $\mathbf{S}\subset \mathbf{V}$ in $\mathcal{G}$, respectively. A graph is weighted if every edge in the graph is associated with a non-zero value. We will work with weighted graphs and treat unweighted graphs as weighted ones whose edge weights are all $1$. A $d$-node weighted graph can be represented by a $d\times d$ matrix $W$, called weighted adjacency matrix, where $W(i,j)$ is $0$ if $X_i\nrightarrow X_j$ and is the weight of $X_i\to X_j$ otherwise. The binary adjacency matrix $A\in \{0,1\}^{d\times d}$ is such that $A(i,j)=1$ if $X_i\to X_j$ in $\cal G$ and $A(i,j)=0$ otherwise. The rank of a weighted graph is defined as the rank of the associated weighted adjacency matrix.
	
	
	A bipartite graph is a graph whose vertex set $\mathbf{V}$ can be partitioned into two disjoint subsets $\mathbf{V}_0$ and $\mathbf{V}_1$, such that the vertices within each subset are not adjacent to one another. $\mathbf{V}_0$ and $\mathbf{V}_1$ are called the parts of the  graph. A matching of a  graph is a subset of its edges where no two of them share a common endpoint. A vertex cover of a graph is a subset of the vertex set where every edge in the graph has at least one endpoint in the subset. The size of a matching (vertex cover) is the number of edges (vertices) in the matching (vertex cover). A maximum matching of a  graph is a matching of the largest possible size and a minimum vertex cover is a vertex cover of the smallest possible size. 
	K{\"o}nig's theorem~\cite{denes1931grafok} states that the size of a minimum vertex cover is equal to the size of a maximum matching in a bipartite graph.
	
	\subsection{Causal Structure Learning and Recent Methods}\label{sec:sec:structure_learning}
	We assume that the data generating procedure follows a structural causal model (SCM). Let  $\mathcal G$ be a DAG. Then an SCM defined with respect to (w.r.t.) $\mathcal G$ is given by: \[X_i = f_i\left(\mathrm{pa}(X_i, {\cal G}), \epsilon_i\right), \quad i=1,2,\dots, d,\]
	where $f_i$ is a deterministic function and $\epsilon_i$'s are jointly independent noises. The SCM induces a marginal distribution $P(X)$ over $X=[X_1, X_2, \cdots, X_d]^T$. In this paper, we assume that there is no latent variable or selection bias. 
	
	The problem of causal structure learning is to infer $\mathcal G$ based on a number of independent and identically distributed (i.i.d.) observations from $P(X)$. In the following, we briefly review the recently developed gradient-based methods that use a smooth characterization of acyclicity of directed graphs, which 
	can be categorized into two classes. 
	
	The first class explicitly associates the target causal model with a weighted adjacency matrix $W$ and then estimate $W$ by solving  optimization problems in the following form:
	\begin{align}
	\label{eq:explicit}
	\min_{W, \phi}~\mathbb E_{P(X)}\,\mathcal S\big(X, h(X;W, \phi)\big), ~~\mathrm{s.t.}~g(W) = 0,
	\end{align}
	where $h:\mathbb R^d\to\mathbb R^d$ is a model function parameterized by $W$ (and possibly other parameters $\phi$) that aims to reconstruct $X$ from $\mathrm{pa}(X_i, {\cal G})$,  $\mathcal S(\cdot, \cdot)$ denotes a score function between the true and reconstructed variables. Here $g(W)\coloneqq \mathrm{tr}\left(e^{W\circ W}\right) - d$, where $\mathrm{tr}(\cdot)$ is the trace function, $\circ$ denotes the element-wise product, and $e^M$ is  the matrix exponential of a square matrix $M$. The constraint was proposed by \cite{zheng2018dags}, which is smooth and holds if and only if $W$ indicates a DAG. Methods in this class include: NOTEARS \cite{zheng2018dags} targeting linear models, and DAG-GNN \cite{Yu19DAGGNN} and the GAE approach \cite{Ng2019GAE} that adopt neural networks to model non-linear relationships.
	Notice that an extra sparsity inducing term may be also added when the causal graph is known to be sparse and that the objective is usually chosen from some well studied score functions (or their variants), e.g.,  penalized maximum likelihood \cite{Chickering2002optimal,van2013ell,loh2014high}.  Further details like parameter optimization techniques can be found in  the cited works and references therein.
		
		The second class uses certain functions, with  parameters denoted by $\theta$, to construct a weighted adjacency matrix $W(\theta)$ (or a binary one $A(\theta)$) to represent the causal structure. These methods can be summarized as:
	\begin{align}\label{eq:implicit}
	\min_{\theta,\,\phi}~\mathbb E_{P(X)}\,\mathcal S\big(X, h(X;W(\theta), \phi)\big),~~\mathrm{s.t.}~g(W(\theta))= 0.
	\end{align}
	For example, GraN-DAG \cite{Lachapelle2019grandag} and NOTEARS-MLP  \cite{Zheng2019learning} respectively use neural network path products and partial derivatives between variables to construct  $W(\theta)$. The binary matrix $A(\theta)$ can be sampled according to some distributions with learnable parameters, as in \cite{Klainathan2018sam,ke2019learning,Ng2019masked,Zhu2020causal,ijcai2021-491}.
	
\begin{remark}  While these methods intend to learn a  DAG, the learned DAG may not be identical to the true one due to Markov equivalence \cite{spirtes2000causation}. For such cases, we may convert the obtained DAG to its completed partially directed acyclic graph (CPDAG) as the final estimate. On the other hand, several forms of SCMs have been shown to be identifiable; if a proper score function is used, then the exact solution to the optimization problem is consistent, i.e., same as the true graph with probability asymptotically approaching $1$. See, e.g., \cite{Shimizu2006lingam,Peters2013identifiability,Peters2014causal,Zhang2009identifiability}, among others. 
\end{remark} 
	
	
\section{Graphical Bounds on Ranks}\label{sec:rank}
	
This section is devoted to relating the rank of a graph to  interpretable graphical conditions, for a better understanding of what kinds of DAGs tend to be low rank. Section~\ref{sec:sec:problem} gives a formal definition of the problem, and Sections~\ref{sec:sec:lower} and \ref{sec:sec:upper} respectively present lower and upper bounds on the graph rank based on some structural priors. If this structural information, such as connectivity, distributions of in-degrees and out-degrees, and an estimate of number of hubs, is accessible,  we can then obtain an estimate of the graph rank. 

\subsection{Problem Formulation}\label{sec:sec:problem}
Consider a DAG ${\cal G}=(\mathbf{V}, \mathbf{E})$ with weighted adjacency matrix $W$ and binary adjacency matrix~$A$. Our goal is to find upper and lower bounds on $\operatorname{rank}(W)$ based on  graphical structure; specifically, we focus on the  weighted adjacency matrices with the same binary adjacency matrix $A$, i.e.,  $\mathcal{W}_A = \{W\in{\mathbb R}^{d\times d}\,;\; \mathrm{sign}(|W|) = A\}$,
	where $\mathrm{sign}(\cdot)$ and $|\cdot|$ are point-wise sign and absolute value functions, respectively. While trivial upper bound $(d-1)$ and lower bound $0$ exist for any DAG, they are generally too loose for practical use. In the following, we  investigate the maximum rank  $\max\{\operatorname{rank}(W); W\in\mathcal{W}_A\}$  and  minimum rank $\min\{\operatorname{rank}(W); W\in\mathcal{W}_A\}$, for tighter upper and lower bounds for any $W \in \cal W_A$. For ease of presentation, we will use $\min\{\operatorname{rank}(\mathcal{W}_A)\}$ and $\max\{\operatorname{rank}(\mathcal{W}_A)\}$ to represent the minimum and maximum ranks, respectively.

To proceed, we introduce two useful graph concepts, namely, \emph{height} and \emph{head-tail vertex cover}.
	
	\begin{definition}[Height]\label{def:height}
		Given a DAG ${\cal G}=(\mathbf{V}, \mathbf{E})$ and a vertex $X_i$, the height of $X_i$, denoted by $l(X_i)$, is the length of the longest directed path starting from $X_i$. The height of $\cal G$, denoted by $l({\cal G})$, is  the length of the longest path in ${\cal G}$.
	\end{definition}
	
	Based on the heights of vertices in $\mathbf{V}$, we can define a weak ordering among the vertices:  $X_i\succ X_j$ if and only if $l(X_i)>l(X_j)$, and $X_i\sim X_j$ if and only if $l(X_i)=l(X_j)$.
	Let $\mathbf{V}_s = \{X_i; l(X_i) = s\}$, $s=0,1,\ldots,l(\cal G)$, and let $\mathbf{V}_{-1}=\varnothing$. It can be verified that: (1) for any given $s\in\{0,1,\ldots,l(\cal G)\}$ and two distinct vertices $X_1, X_2\in \mathbf{V}_s$, $X_1$ and $X_2$ are not adjacent, and (2) for any given $s\in\{1,2,\ldots,l(\cal G)\}$ and $X_i\in\mathbf{V}_s$, there is at least one vertex in $\mathbf{V}_{s-1}$ which is a child of $X_i$. If we denote the induced subgraph of $\cal G$ over $\mathbf{V}_s\cup \mathbf{V}_{s-1}$ by ${\cal G}_{s,s-1}$, then  ${\cal G}_{s,s-1}$ is a bipartite graph with $\mathbf{V}_{s}$ and $\mathbf{V}_{s-1}$ as parts, and singletons in ${\cal G}_{s,s-1}$ (i.e., vertices that are not endpoints of any edge) only appear in $\mathbf{V}_{s-1}$.
	
	\begin{definition}[Head-tail vertex cover]\label{def:pcb}
		Let ${\cal G}=(\mathbf{V}, \mathbf{E})$ be a directed graph and $\mathbf{H}, \mathbf{T}$ be two subsets of $\mathbf{V}$. Then $(\mathbf{H}, \mathbf{T})$ is called a  head-tail vertex cover of $\mathcal{G}$ if every edge in $\mathcal{G}$ has its head vertex in $\mathbf{H}$ or its tail vertex in $\mathbf{T}$. The size of a head-tail vertex cover $(\mathbf{H}, \mathbf{T})$ is defined as $|\mathbf{H}|+|\mathbf{T}|$.
	\end{definition}
	
As an example, consider a directed line graph with $d$ vertices and $d-1$ edges, i.e., $X_1\to X_2\to\cdots\to X_d$. Then $\textbf{H} = \{X_i\}_{i=1}^{d}$ and $\textbf{T}=\varnothing$ form a head-tail vertex cover, with size $d-1$.

\subsection{Lower Bounds}\label{sec:sec:lower}
	
We first study  lower bounds on the rank of a weighted DAG. 
Let $\mathcal{C}(\mathcal{G}_{s,s-1})$ be the set of non-singleton connected components of $\mathcal{G}_{s,s-1}$ and $|\mathcal{C}(\mathcal{G}_{s,s-1})|$ the cardinality. We  have the following lower bounds.
	
	\begin{theorem}\label{thm:lower}
		Let $\cal G$ be a DAG with binary adjacency matrix $A$. Then we have
		\begin{align}
		\min\{\operatorname{rank}(\mathcal{W}_A)\}  \geq 
		\sum\nolimits_{s=1}^{l({\cal G})} |\mathcal{C}(\mathcal{G}_{s,s-1})|  \label{eqn:tighterbd_connected}  \geq 
		l(\cal G). \nonumber
		\end{align}
	\end{theorem}
	
	The complete proofs in this paper are provided in Appendix \ref{app:app:proof}. Theorem~\ref{thm:lower} shows that $\operatorname{rank}(W)$ is greater than or equal to the sum of the number of non-singleton connected components in each $\mathcal{G}_{s,s-1}$. As $\mathcal{G}_{s,s-1}$ has at least one non-singleton connected component, we obtain the second inequality. That is, the rank of a weighted DAG is at least as high as the length of the longest directed path. 	
	
	
Theorem~\ref{thm:lower} also indicates that a sparse graph may have a very high rank. For example, according to Theorem \ref{thm:lower}, the  minimum rank of a directed line graph is $d-1$. As $d-1$ is a trivial upper bound of any DAG, the rank of a directed line graph always equals its number of edges. On the other hand, for some non-sparse graphs, we can assign the edge weights so that the resulting graphs have low ranks. A simple example would be a fully connected directed balanced bipartite graph. A bipartite graph is called balanced if its two parts contain the same number of vertices. The rank of a fully connected balanced bipartite graph with $d$ vertices is $1$ if the edge weights are the same, but the number of edges is $d^2/4$. We remark that to exactly characterize the minimum rank  is still an on-going research problem \cite{HOGBEN20101961}.
	
	\subsection{Upper Bounds}\label{sec:sec:upper}
	We turn to another important issue regarding  the  upper bounds on $\operatorname{rank}(W)$. The next theorem characterizes $\max\{\operatorname{rank}(\mathcal{W}_A)\}$ w.r.t.~graphical terms. 
	
	\begin{theorem}\label{thm:matching}
		Let $\cal G$ be a directed graph with binary adjacency matrix $A$. Then $\max\{\operatorname{rank}(\mathcal{W}_A)\}$ is equal to the minimum size of the head-tail vertex cover of ${\cal G}$. 
	\end{theorem}

	We comment that Theorem~\ref{thm:matching} holds for all directed graphs (not only DAGs), which may be of independent interest in other applications. A head-tail vertex cover of minimum size is called a minimum head-tail vertex cover, which in general is not unique. For a head-tail vertex cover $(\mathbf{H}, \mathbf{T})$, the vertices in $\mathbf{H}$ cover all the edges pointing towards these vertices  while the vertices in $\mathbf{T}$ cover the edges pointing away. A head-tail cover of a relatively small size then indicates the presence of hubs, that is, vertices with relatively high in-degrees or  out-degrees. Therefore, Theorem~\ref{thm:matching} suggests that the maximum rank of a weighted DAG is highly related to the presence of hubs: a DAG with many hubs tends to have a low rank. Intuitively, a hub of high in-degree (resp.~out-degree) is a common effect (resp.~cause) of a number of direct causes (resp.~effect variables), comprising many V-structures (resp.~inverted V-structures). 

Such features are frequently encountered in real-world causal networks, and the existence of such features may be part of domain knowledge. For instance, it was claimed in \cite{barabasi2004network} that some protein networks, which are directed and acyclic due to irreversible reactions, are the results of growth processes and preferential attachments,  due to gene duplication. Thus, many gene and protein networks tend to have hubs of high out-degree. Another example is from a financial network that describes the direction of risk contagion: the insurance companies take risks from many other financial institutions, and thus are likely to have high in-degrees \cite{GAO20132965}.

\begin{figure}[t]
		\centering
		\subfigure{
\includegraphics[width=0.46\textwidth]{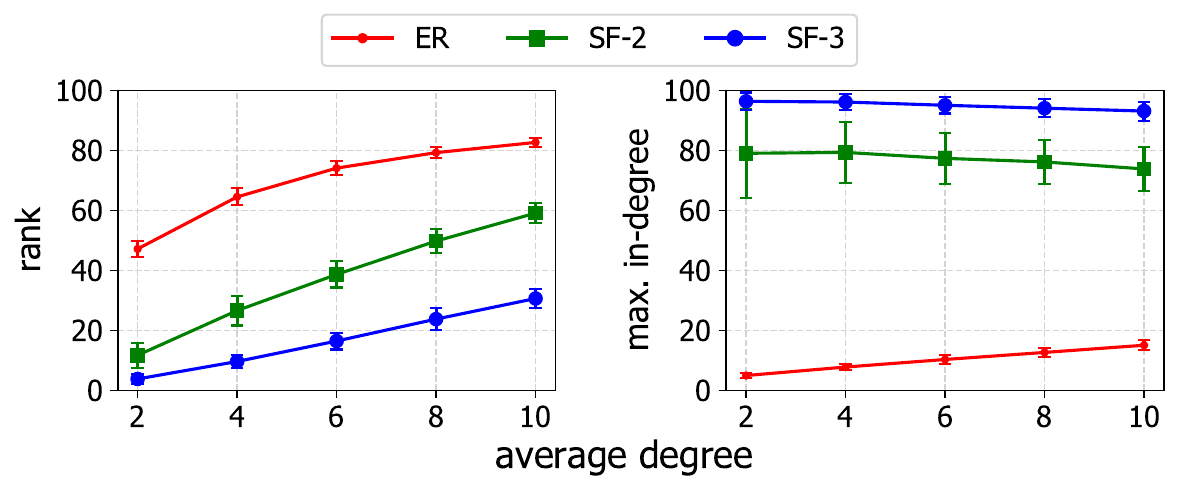}
		}
		\vspace{-0.5em}
		\caption{Graph ranks ({\bf left}) and maximum in-degrees ({\bf right}) versus average degrees with $100$-node ER and SF random graphs.}  
		\label{fig:sfer}
 		\vspace{-1em}
	\end{figure}

	In addition, the frequently encountered scale-free (SF) graphs have hubs and tend to be low rank~\cite{Guelzim2002TopologicalAC, barabasi2004network, hartemink2005reverse, PhysRevLett.94.018102, GAO20132965, Ramsey2017million}. A scale-free graph is one whose distribution of degree $k$ follows a power law: $P(k) \sim k^{-\gamma}$, where $\gamma$ is the power parameter typically within $[2,3]$ and $P(k)$ denotes the fraction of nodes with degree $k$ \cite{Nikolova2012Parallel}. Empirically, the ranks of scale-free graphs are relatively low, especially in comparison to Erd{\"o}s-R{\'e}nyi (ER) random graphs~\cite{mihail2002eigenvalue}. Figure~\ref{fig:sfer} provides a simulated example where $\gamma$ is chosen from $\{2,3\}$ and each reported value is over $500$ random runs.  As graph becomes denser, the graph rank also increases. However, for scale-free graphs, their ranks are much lower than those of ER graphs. We also show maximum in-degrees of the simulated graphs in Figure~\ref{fig:sfer}. As discussed in Section~\ref{sec:intro},  the maximum in-degree has a large effect on the performance (e.g., search space and sample complexity) of many causal structure learning methods and scale-free graphs were conjectured as a hard case in \cite{Ramsey2017million}.  
	
	Next, we report some simpler upper bounds. 
	
	\begin{theorem}\label{thm:upper-loose}
		{Let $\cal G$ be a DAG with binary adjacency matrix $A$. Denote by $\mathbf{V}_{\mathrm{ch}}$ the set of vertices that are children of some vertices  and by $\mathbf{V}_{\mathrm{pa}}$ those that are parents of some vertices.  Then we have}
		\begin{equation}
		\max\{\operatorname{rank}({\mathcal W}_A)\}\hspace{-2pt}\leq\hspace{-2pt}
		\left\{
		\begin{aligned}
		&\hspace{-4pt}\sum\nolimits_{s=1}^{l}\hspace{-2pt} \mathrm{min}\hspace{-1pt}\left(|\mathbf{V}_s|,\hspace{-1pt} |\mathrm{ch}(\mathbf{V}_s)| \right)\hspace{-2pt}\leq\hspace{-2pt} |\mathbf{V}_{\mathrm{pa}}|,\nonumber\\
		&\hspace{-4pt}\sum\nolimits_{s=0}^{l-1}\hspace{-2pt} \mathrm{min}\hspace{-1pt} \left(|\mathbf{V}_s|,\hspace{-1pt} |\mathrm{pa}(\mathbf{V}_s)|\right)\hspace{-2pt} \leq\hspace{-2pt} |\mathbf{V}_{\mathrm{ch}}|,\nonumber\\
		&\hspace{-2pt}|\mathbf{V}|-\mathrm{max}\{|\mathbf{V}_s|;0\leq s\leq {l({\cal G})}\}.\nonumber
		\end{aligned}
		\right.
		\end{equation} 
	\end{theorem}
	
	
As $\mathbf{V}_{\mathrm{ch}}$ and $\mathbf{V}_{\mathrm{pa}}$ respectively represent the sets of non-root and non-leaf nodes, the first two bounds in Theorem~\ref{thm:upper-loose} indicate that the maximum rank is bounded from above by the number of non-root nodes and also the number of non-leaf nodes. These simple bounds are useful when there is background knowledge indicating that many variables are roots or are leaves in the underlying causal network, e.g., the \emph{ecoli70} \cite{schafer2005shrinkage} and \emph{magic-niab} \cite{scutari2014multiple} networks. 	The last inequality is a generalization of the first two, which implies that the  rank is likely to be low if most vertices have the same height. 
	
	\begin{theorem}\label{thm:upper-undirected}
		Let $\cal G$ be a DAG with binary adjacency matrix $A$. Denote by $\mathrm{skel(A)}$ and $\mathrm{moral(A)}$  the binary adjacency matrices of the skeleton and moral graphs of $\cal G$, respectively. Then we have  
		\begin{equation}
		\begin{aligned}
		\max\{\mathrm{rank}(\mathcal{W}_A) \}\nonumber \leq &  \max\{\mathrm{rank}(W);\, \mathrm{sign}(|W|)=\mathrm{skel}(A)\} \nonumber\\
		\leq & \max\{\mathrm{rank}(W);\,\mathrm{sign}(|W|)= \mathrm{moral}(A) \}.
		\end{aligned}
		\end{equation}
	\end{theorem}
	
	The skeleton of a DAG is the undirected graph obtained by removing all the arrowheads, and the moral graph is the undirected graph where two vertices are adjacent if they are adjacent  or if they share a common child in the DAG.  This theorem may be useful when the skeleton or the moral graph can be easily estimated from data, e.g., in a linear Gaussian model where the zero-nonzero pattern of the precision matrix indicates the moral graph.
	
			\begin{algorithm}[t]
		\caption{Generating random DAGs}
		\label{algo:app:randDAG}
		\begin{algorithmic}[1]
			\REQUIRE
			Number of nodes $d$, average degree $k$, and rank $r$.
			\ENSURE
			A randomly sampled DAG with the number of nodes $d$, average degree $k$, and rank $r$.
			\STATE {Set $M=\,$ empty graph, $M_p=\varnothing$,  $R=\{(i, j);\; i<j, \,i,j=1,2,...,d\}$, and $p=k/(d-1)$.}
			\STATE {Sample a number $N\sim \mathcal{B}(d(d-1)/2, p)$, where $\mathcal{B}(n, p)$ is a binomial distribution with parameters $n$ and $p$.}
			\IF {$N<r$}
			\STATE {return FAIL}
			\ENDIF
			\STATE {Sample $r$ indices from $1,\ldots,{d-1}$ and store them in $M_p$ in descending order.}
			\FOR{each $i$ in $M_p$}
			\STATE{Sample an index $j$ from $i+1$ to $d$, add edge $(i,j)$ to $M$, and remove $(i, j)$ from $R$.}
			\ENDFOR
			\WHILE{$R\neq\varnothing$ and $|M|<N$}
			\STATE{Sample an edge $(i, j)$ from $R$ and remove it from $R$.}
			\STATE{Add $(i, j)$ to $M$ if it does not change the size of the minimum head-tail vertex cover of $M$.}
			\ENDWHILE
			\IF {$|M|<N$}
		\STATE {return FAIL}
		\ENDIF
		\STATE {Permute randomly the vertex indexes in $M$.}
		\STATE {Assign the edge weights of $M$ randomly according to a continuous distribution.}
		\STATE {return $M$}
	    \end{algorithmic}
    \end{algorithm}

\subsection{Generating Low Rank DAGs}
In this section, we discuss another useful result from
Theorem~\ref{thm:matching}, i.e., to generate a random DAG with a given rank $r$ and a properly specified average degree $k$, as shown in Algorithm~\ref{algo:app:randDAG}. The first part of Algorithm \ref{algo:app:randDAG}f is to sample a number $N$, representing the total number of edges, from a binomial distribution $\mathcal{B}(d(d-1)/2, p)$ where $p=k/(d-1)$. If $N<r$, Algorithm~\ref{algo:app:randDAG} would return FAIL since a graph with $N<r$ edges could never have rank~$r$. 
Otherwise, an initial graph with $r$ edges and rank $r$ is sampled, which can be done by choosing the edges such that no two of them share the same head point or the same tail point, i.e., each row and each column of the corresponding adjacency matrix have at most one non-zero entry. Then Algorithm~\ref{algo:app:randDAG} samples an edge from the set $R$ that contains all possible edges and checks whether adding this edge to the graph changes the size of the minimum head-tail vertex cover. If not, the edge will be added to the graph; otherwise, it will be removed from $R$. This is because if a graph $\cal G$ is a super-graph of another graph $\cal H$, then the size of the minimum head-tail cover of $\cal G$ is no less than that of $\cal H$. We repeat the above sampling procedure until there is no edge in $R$ or the number of edges in the resulting graph reaches $N$. If the latter happens, the algorithm will return the generated graph; otherwise, it will return FAIL.

Note that the algorithm may not return a valid graph if the desired number $N$ of edges cannot be reached. This could happen if the input rank is too low while the input average degree is too high. Fortunately, with our experiment settings, we find it rare for Algorithm \ref{algo:app:randDAG} to fail to return a desired graph. In fact, there exist some connections between the maximum rank and the number of edges in the graph, according to Theorem~\ref{thm:matching}. Intuitively,  if the graph is dense, then we need more vertices to cover all the edges. Thus, the size of the minimum head-tail vertex cover tends to be large. Explicitly providing a formula to characterize these two graph parameters is an interesting future direction. 

\section{Exploiting Low Rank Assumption in Causal Structure Learning}\label{sec:learning}
	
Thanks to the studies of low-rankness in various contexts, we can adapt  gradient-based causal structure learning methods with little extra effort based on existing techniques. We remark that our adaptations with the low rank assumption are not restricted to a particular DAG learning algorithm; other gradient-based  methods may incorporate one of the modifications too, if they have similar formulations and optimization procedures. 

	{\bf Matrix Factorization} \quad Since the weighted adjacency matrix $W$ is explicitly optimized in the first class of methods, we can  apply the matrix factorization technique (e.g.~\cite{Koren2009Matrix, Nathan03ICML, Chi2019Nonconvex}). Specifically, with an estimate $\hat r$ for the graph rank, we can factorize $W$ as $W=UV^T$ with $U, V\in \mathbb{R}^{d\times \hat{r}}$. Problem~(\ref{eq:explicit}) is then to optimize $U$ and $V$ that minimizes the score function under the DAG constraint, and has the same solution $W$ (obtained from $UV^T$) as the original one if $\hat r$ is greater than or equal to the true rank. Furthermore, if $\hat{r}\ll d$, we  have a much reduced number of parameters to optimize.

	{\bf Nuclear Norm}\quad For the second class of methods, the adjacency matrix $W(\theta)$ is not an explicit parameter to be optimized. In such a case, we can  add a nuclear norm term $\lambda\|W(\theta)\|_*$, with $\lambda>0$ being a tuning parameter, to the objective to induce low-rankness (e.g.,~\cite{Martin10icml, Gu_2014_CVPR, Fazel2001minimization, Hu2013Fast, gu2017weighted}). Notice that this approach is also feasible for the first class of methods, but we find that it does not work as well as the matrix factorization approach, possibly due to the singular value decomposition to compute the (sub-)gradient w.r.t.~$W$ at each optimization step.

Both the matrix factorization and  nuclear norm approaches enjoy favorable theoretical guarantee and empirical performance in many  low rank applications. Similar to existing cases, the optimization procedures in these structure learning methods are gradient-based and can easily handle the two adaptations. Below we describe details about the optimization procedures. It would be beneficial to tailor the optimization for causal structure learning, but this is beyond the scope of the present paper.  

	\subsection{Optimization}\label{app:app:solve}
	For this part, we consider a dataset consisting of $n$ i.i.d.~observations from $P(X)$ and consequently the expectations  in Problems~(\ref{eq:explicit}) and (\ref{eq:implicit}) are replaced by empirical means. Denote the design matrix by $\mathbf{X}\in \mathbb{R}^{n\times d}$, where each row of $\mathbf{X}$ corresponds to an observation   and each column represents a variable. Here we use NOTEARS \cite{zheng2018dags} (and similarly GOLEM \cite{ng2020role})  and Gran-DAG \cite{Lachapelle2019grandag}  from each class as examples and will describe their low rank versions in the following. Other gradient-based methods and their optimization procedures can be similarly modified to incorporate the low rank assumption.
	
	\subsubsection{NOTEARS with Low Rank Assumption}
	Following Section \ref{sec:learning} as well as the work of NOTEARS \cite{zheng2018dags}, the optimization problem  can be written as
	\begin{equation}
	\label{eqn:low-rank}
	\begin{aligned}
	\min_{U,V} \frac{1}{2 n} \left\|\mathbf{X}-\mathbf{X}UV^{T} \right\|_{F}^{2}, ~~\mathrm{s.t.}~g(UV^{T}) = 0,
	\end{aligned}
	\end{equation}
	where $U, V\in \mathbb{R}^{d\times \hat{r}}$, $\|\cdot\|_{F}$ denotes the Frobenius norm, and $g(\cdot)$ was introduced in Section~\ref{sec:sec:structure_learning}  to induce a DAG. 
	This problem can then be solved by standard numeric optimization methods such as the augmented Lagrangian method~\cite{Bertsekas/99}. In particular,  the augmented Lagrangian is given by 
	\[L_{\rho}(U, V, \alpha) \hspace{-2pt}=\hspace{-2pt} \frac{1}{2 n} \left\|\mathbf{X}\hspace{-2pt}-\hspace{-2pt}\mathbf{X}UV^{T} \right\|_{F}^{2} +\alpha g(UV^{T})+\frac{\rho}{2}\left|g(UV^{T})\right|^{2}\hspace{-2pt},\]
	where
	$\alpha$ is the Lagrange multiplier and $\rho>0$ is the penalty parameter. The optimization procedure is summarized in Algorithm~\ref{algo:app:raopt}, similar to \cite[Algorithm~1]{zheng2018dags}.

		\begin{algorithm}[!t]
		\caption{Optimization procedure for NOTEARS-low-rank}
		\label{algo:app:raopt}
		\begin{algorithmic}[1]
			\REQUIRE
			Design matrix $\mathbf{X}$, starting point $(U_0, V_0, \alpha_0)$, rate $c\in(0, 1)$, tolerance $\epsilon>0$, and threshold $w>0$.
			\ENSURE
			Locally optimal parameter $W^{*}$.
			\FOR {$t=1,2,\ldots$}
			\STATE {$U_{t+1}, V_{t+1} \leftarrow \mathrm{arg}\;\mathrm{min}_{U, V}\;L_{\rho}(U, V, \alpha_t)$ with $\rho$ such that $g(U_{t+1} V_{t+1}^{T})<c g(U_{t} V_{t}^{T})$.}
			\STATE {$\alpha_{t+1} \leftarrow \alpha_t+\rho g(U_{t+1} V_{t+1}^{T})$.} 
			\IF {$g(U_{t+1} V_{t+1}^{T})<\epsilon$}
			\STATE {Set $U^{*}=U_{t+1}$ and $V^{*}=V_{t+1}$.}
		\STATE {\textbf{break}}
			\ENDIF
			\ENDFOR
			\STATE {Set $W^{*}=(U^{*}V^{*T})\circ \mathbf{1}(|U^{*}V^{*T}|>w)$.}
			\STATE {return $W^{*}$}
	\end{algorithmic}
	\end{algorithm}
The unconstrained subproblem in Step~2 can be solved by existing optimization methods such as  L-BFGS and Newton conjugate gradient method. Notice that the DAG constraint may not be satisfied exactly using iterative numeric methods, so it is a common practice to pick a small tolerance, followed by a thresholding procedure on the estimated entries  to obtain an exact DAG. This heuristic is made possible by virtue of the DAG penalty term, which pushes
the cycle-inducing edges to small values. In our implementation,  we choose $U_0$ and $V_0$ to be the first $\hat{r}$ columns of the $d\times d$ identity matrices. Other parameter choices are: $\alpha_0=0$, $c=0.25$, $\epsilon=10^{-6}$, and $w=0.3$, similar to those used in related methods on the same datasets (e.g., \cite{zheng2018dags, Yu19DAGGNN,Zhu2020causal}).  The chosen threshold $w=0.3$ works well  in our experiments as well as in the experiments of related works that use the same data model. In case the thresholded matrix is not a DAG, one can further increase the threshold until the resulting matrix corresponds to a DAG.
	
After obtaining $W^*$, we apply an additional pruning step: we use linear regression to refit the dataset based on the structure indicated by $W^*$ and then  apply another thresholding (with the same threshold as the previous thresholding procedure) to the refitted weighted adjacency matrix. The additional pruning step is  applied to original NOTEARS, which also improves its performance with lower Structural Hamming Distance (SHD), particularly for large and dense graphs. 

 Note that we do not include the $\ell_1$ penalty term w.r.t.~$UV^T$ in the above version, for the following reasons: (1) the  thresholding procedure can also control false discoveries; (2) we consider relatively sufficient data for the experiments and NOTEARS with thresholding has been shown in \cite{zheng2018dags} to perform consistently well even when the graph is sparse; (3) we are more concerned with relatively large and dense graphs, so a sparsity assumption may be harmful, as shown also by \cite{zheng2018dags}. Nevertheless,  we include NOTEARS with $\ell_1$ penalty in the first and last experiments in Section~\ref{sec:exp} for more information regarding the role of sparsity, which verifies  the advantage of NOTEARS without $\ell_1$ penalty over that with $\ell_1$ penalty for relatively dense graphs.

\subsubsection{GOLEM with Low Rank Assumption}	
Instead of the hard acyclicity constraint, GOLEM \cite{ng2020role} formulates causal structure learning with linear data models with a soft one, in the following form:
\begin{align}
\min_W \frac{1}{2n} \log\left(\left\|\mathbf{X}-\mathbf{X}W \right\|_{F}^{2}\right) + \lambda_1\|W\|_1+\lambda_2 \mathrm{tr}\left(e^{W \circ W}\right). \nonumber
\end{align}
The authors show that such a formulation works well on linear Gaussian data models with equal noise variances, even for very large problems. As such, GOLEM and its low rank version (as described below) will be used for the large problems with $800$ nodes in our experiment in Section~\ref{sec:sec:ex_lgsem}. For linear Gaussian models with non-equal noise variances, the authors replace $\log(\left\|\mathbf{X}-\mathbf{X}W \right\|_{F}^{2})$ 
with 
$\sum_{i=1}^d(\log(\sum_{l=1}^n(X_i^{(l)}-W_i^TX^{(l)})^2))$,
where $W_i$ is the $i$-th column of $W$, $X^{(l)}$ is the $l$-th observation and $X_i^{(l)}$ is the $i$-th entry in the $l$-th observation. 

Similar to NOTEARS, GOLEM-low-rank is then given by
	\begin{align}
	\min_{U,V} \frac{1}{2n} \left\|\mathbf{X}\hspace{-2pt}-\hspace{-2pt}\mathbf{X}UV^{T} \right\|_{F}^{2}\hspace{-2pt} + \hspace{-2pt} \lambda_1\|UV^T\|_1\hspace{-2pt}+\hspace{-2pt}\lambda_2\left(e^{(UV^{T}) \circ (UV^{T})}\right),\nonumber
	\end{align}
which can be solved in a similar manner to Algorithm~\ref{algo:app:raopt}. The estimate $W^*$ is then calculated by the solutions $U^*$ and $V^*$, i.e., $W^*=U^*V^{*T}$. The same post-processing procedure of original GOLEM is then applied. Here the choices of $\lambda_1$, $\lambda_2$, and the threshold are made in the same way as in \cite{ng2020role}.

\subsubsection{GraN-DAG with low rank assumption}
	We next consider a low rank version of GraN-DAG. The optimization problem can be written as
	\begin{equation}
	\begin{aligned}
	\min_\theta&-\hspace{-2pt}\frac{1}{n}\hspace{-1pt}\sum_{l=1}^{n}\hspace{-2pt}\sum_{i=1}^{d}\hspace{-1pt} \log p\left(X_i^{(l)}\mid {\rm pa}(X_i,\hspace{-2pt} W(\theta))^{(l)};\theta\right)\hspace{-2pt}+\hspace{-2pt}\lambda\|W{(\theta)}\|_{*}\nonumber\\
	\mathrm{s.t.}~&g(W{(\theta)})= 0, \nonumber
	\end{aligned}
	\end{equation}
	where $X_i^{(l)}$ is the $l$-th sample of variable $X_i$ and $ {\rm pa}(X_i, W(\theta))^{(l)}$ means the $l$-th sample of $X_i$'s parents  indicated by the adjacency matrix $W{(\theta)}$. Here, $\theta$ denotes the parameters of neural networks and $W{(\theta)}$ with non-negative entries is obtained from the neural network path products. 
	
	The above problem can be solved similarly using augmented Lagrangian. The procedure is similar to Algorithm \ref{algo:app:raopt} and is the same to that used by GraN-DAG, with slight modifications: (1) the subproblem in Step 2 is approximately solved using first-order methods;  (2) the thresholding at Step 9 is replaced by a variable selection method proposed by \cite{Buhlmann2014cam}. The same variable selection or pruning method is adopted by two other benchmark methods CAM and NOTEARS-MLP in our experiment. Please refer to \cite{Lachapelle2019grandag} and \cite{Buhlmann2014cam} for further details.

\subsection{Hyperparameters}
In practice, we need to pick hyperparameters $\hat r$ and $\lambda$ for the low rank modifications. For the first class of methods, if we know that an estimate $\hat r$ is identical or close to the true rank, then we can directly plug it into the factorized method. While this case may be rare in practice, we use it as  a sanity check of the low rank methods (cf.~Section~\ref{sec:sec:ex_lgsem}). Perhaps it is more often to know only a range of possible ranks based on certain prior information, e.g., some structural information in Theorems~1--4, or when the underlying DAG is scale-free and  a rough estimate of average degree and power is available (as experimentally shown in Section~\ref{sec:scale_free_exp}). Similar to selecting other hyperparameters, we may determine $\hat{r}$ assisted by a validation dataset (or by cross-validation if the observed dataset is not sufficiently large), or try different choices and then apply traditional score-based method where the search space is restricted to the resulting DAGs. As shown in Section~\ref{sec:sentivity}, the modified algorithm performs consistently well over a large range of rank parameters, so we may evenly pick $5$--$10$ choices within the range; this strategy is further verified by our experiments in Section~\ref{sec:exp}. 

For the second class of methods with a nuclear norm regularization, we may also apply the validation approach with several penalty weights. While the graph rank does not directly indicate an appropriate choice and a fixed set of candidate weights may not work well for all problems, the existing methods for selecting similar hyperparamters (like the weights of $\ell_1$ or $\ell_2$ penalties) can be very useful here. In our experience, choosing from $\{0.1, 0.3, 0.5, 0.8, 1, 2, 5\}$ works reasonably well, as shown in Section~\ref{nonlinearSEMs}. 

\begin{remark}
{When it is not certain that the underlying DAG is low rank, we can also include the original algorithms in the validation approach.  Section~\ref{nonlinearSEMs} empirically verifies this strategy with non-linear SCMs when the graphs to be learned are not low rank.}
\end{remark}

\section{Experiments}\label{sec:exp}
		
	This section reports empirical results of the low rank adaptations of existing methods, compared with their original versions. We focus on NOTEARS \cite{zheng2018dags} and GOLEM \cite{ng2020role} for linear data models by adopting the matrix factorization approach, denoted respectively as NOTEARS-low-rank and GOLEM-low-rank, and use the nuclear norm approach in combination with GraN-DAG \cite{Lachapelle2019grandag} for a non-linear SCM. These modifications have be described in  Section~\ref{app:app:solve}, and an implementation has been released in the \texttt{gCastle} package \cite{gcastle}.\footnote{{https://github.com/huawei-noah/trustworthyAI}} For more information, we also include several benchmark methods: fast GES \cite{Ramsey2017million}, PC \cite{spirtes2000causation}, MMHC \cite{Tsamardinos2006max}, the polynomial-time method from \cite{Ghoshal2018learning},  DirectLiNGAM \cite{Shimizu2011directlingam} and ICA-LiNGAM \cite{Shimizu2006lingam} that are specifically designed with non-Gaussian noises, for linear cases; and DAG-GNN \cite{Yu19DAGGNN}, NOTEARS-MLP \cite{Zheng2019learning}, and CAM \cite{Buhlmann2014cam} for the non-linear SCMs. Here Fast GES and PC are called from {\tt py-causal} package\footnote{{https://github.com/bd2kccd/py-causal}} and MMHC is called from {\tt bnlearn} package.\footnote{{https://CRAN.R-project.org/package=bnlearn}} The rest methods have available implementations released in the respective papers and we  use default hyperparameters unless otherwise stated.

We consider randomly sampled DAGs with specified ranks generated by Algorithm~\ref{algo:app:randDAG}, scale-free graphs, and two real networks. For the linear data model of the following form, 
\begin{equation}\label{eq:lgmodel}
	X_i = \sum_{X_j\in \mathrm{pa}(X_i, {\mathcal G})} W(j,i)X_j + \epsilon_i, \quad i=1,2,\ldots,d,
\end{equation}
	we follow existing works (such as NOTEARS and DAG-GNN) to draw the weights form $\operatorname{Unif}([-2,-0.5]\cup[0.5, 2])$, and assume $\epsilon_i$'s to be jointly independent noises with three cases: (1) standard Gaussian distributions; (2) standard exponential distributions; and (3) zero-mean Gaussian distributions with variances drawn from $\operatorname{Unif}([0.5, 2])$. For non-linear SCM, we consider the following model which are also used in the works of CAM and NOTEARS-MLP.
	\begin{equation}\label{eq:gpmodel}
	X_i = f_i(\mathrm{pa}(X_i, {\mathcal G})) + \epsilon_i, \quad i=1,2,\ldots,d,
	\end{equation}
	where $\epsilon_i$'s are jointly independent standard Gaussian noises and $f_i$'s are functions sampled from Gaussian processes with RBF kernel of bandwidth one.  Below we mainly report structural hamming distance (SHD) (or SHD-CPDAG, the SHD between CPDAGs of estimated graph and ground truth,  for non-identifiable cases), which takes into account both false positives and false negatives, and a smaller SHD indicates a better estimate \cite{Tsamardinos2006max}. Some more  evaluation metrics are provided in Appendix~\ref{app:detailed_result}.

\subsection{Linear Data Models with Rank-Specified Graphs}\label{sec:sec:ex_lgsem}  
We use linear data models on rank-specified  graphs,  with number of nodes $d\in\{100, 300\}$, rank $r=\lceil 0.1d \rceil$, and average degree $k\in \{2,4,6,8\}$. This experiment serves as a sanity check of the  low rank approach---the true rank is assumed to be known and is used as the rank parameter $\hat r$ in {NOTEARS-low-rank} and {GOLEM-low-rank}. From each data model, we then generate $n=3,000$ observations. We repeat $20$ times over different seeds for each experiment setting. The experiments are run on a Linux workstation with 64-core Intel Xeon 3.20GHz CPU, Nvidia Quadro RTX 6000 GPU, and 128GB RAM.

\begin{figure}[t!]
	\centering
	\addtocounter{subfigure}{-1}
	\subfigure{
		\begin{minipage}[t]{0.45\textwidth}
			\centering
			\includegraphics[width=\textwidth]{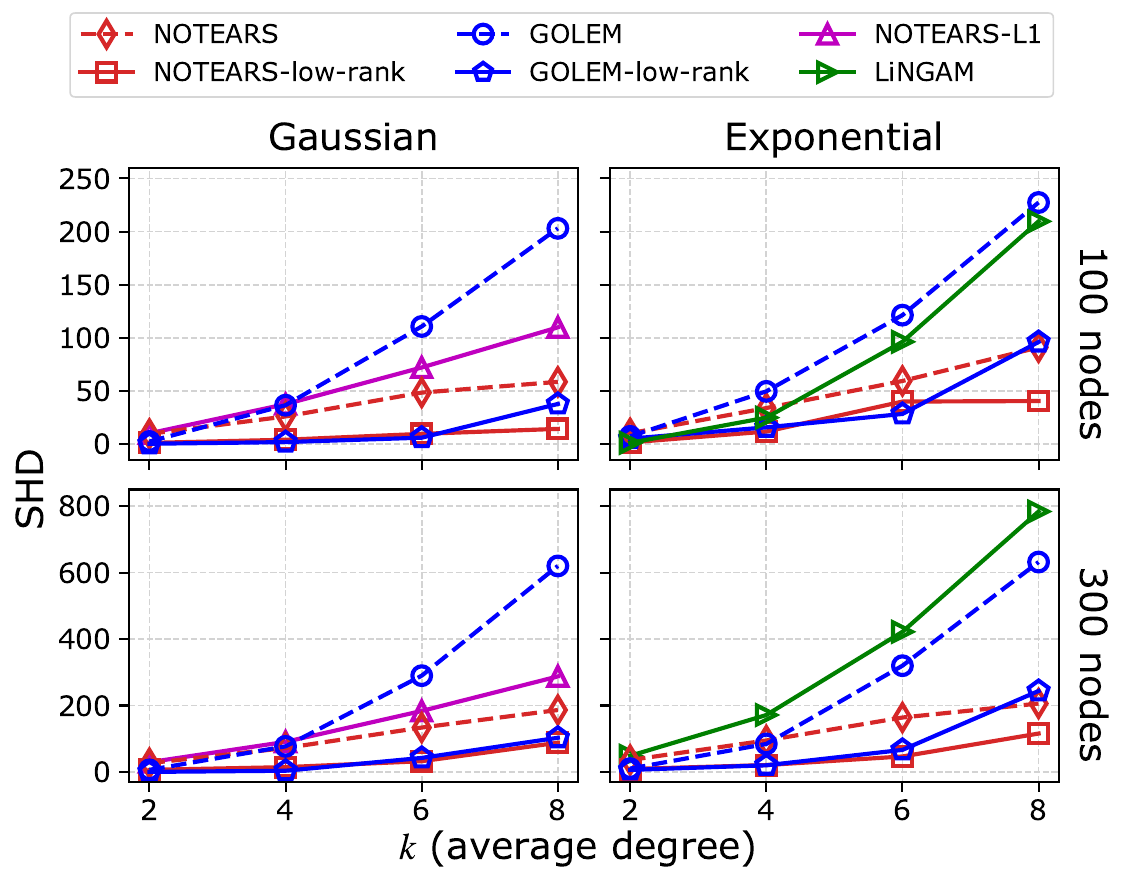}
		\end{minipage}%
	}
\vspace{-0.5em}
	\caption{Average SHDs on rank-specified graphs.} 
	\label{fig:linear_sems}
\vspace{-1em}		
\end{figure}

{\bf Identifiable Case}\quad We first consider additive noises following standard Gaussian, which is known to be identifiable \cite{Peters2013identifiability}. For a better visualization, Figure~\ref{fig:linear_sems} only reports the average SHDs, while the true positive rate, false discovery rate, and running time are left to Appendix~\ref{app:detailed_result}.  We observe that the performance of NOTEARS and GOLEM degrades when the graph becomes dense and the maximum in-degree increases; in particular, the maximum in-degrees are $\{11.0\pm1.6, 22.0\pm2.2, 33.9\pm4.1, 44.7\pm6.3\}$ for $100$-node graphs with average degrees $\{2,4,6,8\}$, respectively. The low rank versions can greatly improve the performance of original algorithms, reducing SHDs by at least a half. The fast GES, PC and MMHC has  much higher SHDs. For the polynomial-time method in \cite{Ghoshal2018learning}, it has a similar performance as NOTEARS-low-rank and GOLEM-low-rank with sparse graphs of degree $2$, but degrades much when graph becomes denser; specifically, for $100$-node graphs, its SHDs are $1.25\pm2.4$ and $307.25\pm52.75$ for degrees $2$ and $8$, respectively. More detailed results are provided in Table~\ref{tab:detail} in the appendix.

For more information regarding the role of sparsity, we include NOTEARS with an $\ell_1$ penalty, denoted as NOTEARS-L1, where the $\ell_1$ penalty weight is chosen from $\{0.01, 0.02, 0.05, 0.1, 0.2, 0.5\}$. Instead of relying on an additional validation dataset, we treat NOTEARS-L1 favorably by picking the lowest SHD obtained from different weights for each dataset. Yet NOTEARS-L1 is slightly better than NOTEARS when the average degree is $2$, and is largely outperformed with relatively dense graphs. This observation was also reported in \cite{zheng2018dags}. We conjecture that it is because our experiments consider relatively sufficient data and dense graphs. The utility of low rank adaptations for relatively dense graphs is also verified with additive exponential noises, where we include  DirectLiNGAM for $100$-node graphs and ICA-LiNGAM for $300$-node graphs (because DirectLiNGAM is too slow for the latter case).


{\bf Non-identifiable Case} \quad We proceed to non-identifiable linear data models with zero-mean Gaussian noises whose variances are  drawn from $\mathrm{Unif}([0.5,2])$. We adopt a version of GOLEM that is specifically developed for non-equal variances and use SHD-CPDAG as the metric. We report the average SHD-CPDAGs for $100$-node graphs with $r=10$ and average degrees $k\in\{2,4,6,8\}$, which are $\{33.5\pm29.1, 144.2\pm32.4,221.9\pm 47.7, 283.7\pm64.1\}$ for GOLEM and $\{3.9\pm5.9, 38.0\pm 40.5,116.5\pm73.8,200.9 \pm 79.1\}$ for GOLEM-low-rank. This again verifies the utility of a low rank adaptation.

\begin{figure}[t!]
		\begin{center}
			\includegraphics[width=0.27\textwidth]{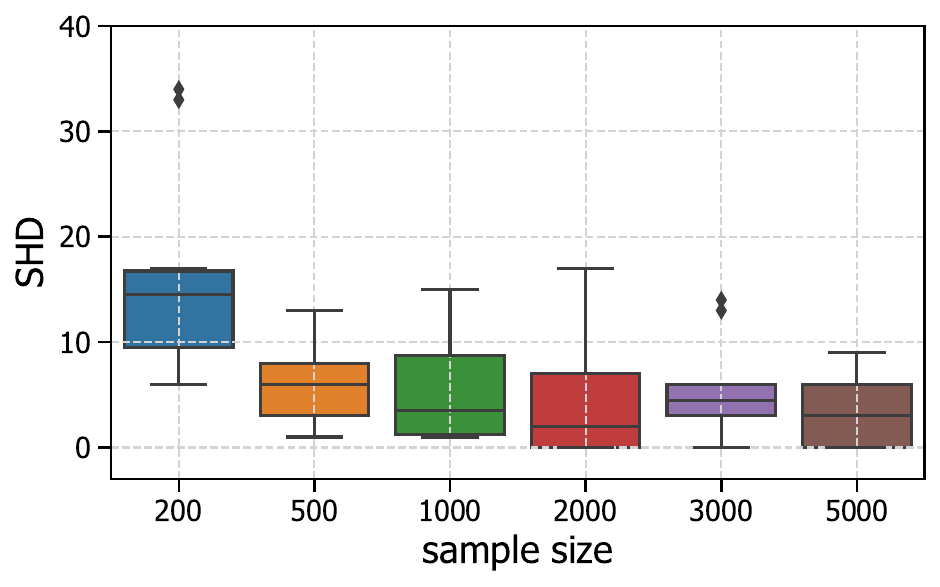}
		\end{center}
		\vspace{-0.5em}
\caption{Different sample sizes.}
 				\label{fig:consistent}
 	\vspace{-1em}
\end{figure}

{\bf Different Sample Size} \quad We conduct an empirical analysis with different sample sizes. Again, we use rank-specified random graphs (sampled according to Algorithm~\ref{algo:app:randDAG}) with $d = 100$ nodes, degree $k=8$, rank $r=10$, and linear Gaussian SEMs. We also assume that the true rank is known. We fix the rank parameter $\hat{r}=10$ and use different sample sizes ranging from $200$ to $5,000$. As reported in Figure~\ref{fig:consistent}, NOTEARS-low-rank performs reasonably well  when  the sample size is small and tends to have a better performance with more samples.

{\bf Higher Rank}\quad This experiment considers graphs of higher ranks. We use rank-specified random graphs with $d = 100$ nodes and rank $r\in\{30, 35, 40 ,45, 50\}$ on linear Gaussian data models. The results are shown in the left and middle panels of Figure~\ref{fig:other_sems}, with average degrees $2$ and $8$, respectively. We observe that when the rank of the underlying graph becomes higher, the advantage of  NOTEARS-low-rank over NOTEARS decreases. Nonetheless, NOTEARS-low-rank with rank $r=50$ is still comparable to NOTEARS, and has a lower average SHD  after removing outlier SHDs using  the interquartile range rule. A possible reason is that the maximum in-degree of the rank-specified graphs, given a fixed average degree, tend to decrease when the rank becomes higher; in particular, for average degree  $8$, the maximum in-degrees are $\{22.6\pm2.2, 19.7\pm3.4, 17.8\pm2.6, 17.5\pm2.4, 14.6\pm1.6\}$ for ranks $\{30, 35,40,45,50\}$, respectively. We then increase the average degree to $14$ and $20$, while fixing rank $r=50$;  the maximum in-degrees become $\{4.4\pm0.8,14.6\pm1.6,25.4\pm2.7,34.6\pm3.2\}$ for average degrees $\{2,8,14,20\}$, respectively. As shown in the right panel of Figure~\ref{fig:other_sems}, both NOTEARS and NOTEARS-low-rank have increasing SHDs, as the graphs are of both high  in-degrees and high ranks.

\setcounter{figure}{2}
\begin{figure*}[t!]
 	\centering
 			\subfigure{
 			\begin{minipage}[t]{0.66\textwidth}
 				\centering	\includegraphics[width=\textwidth]{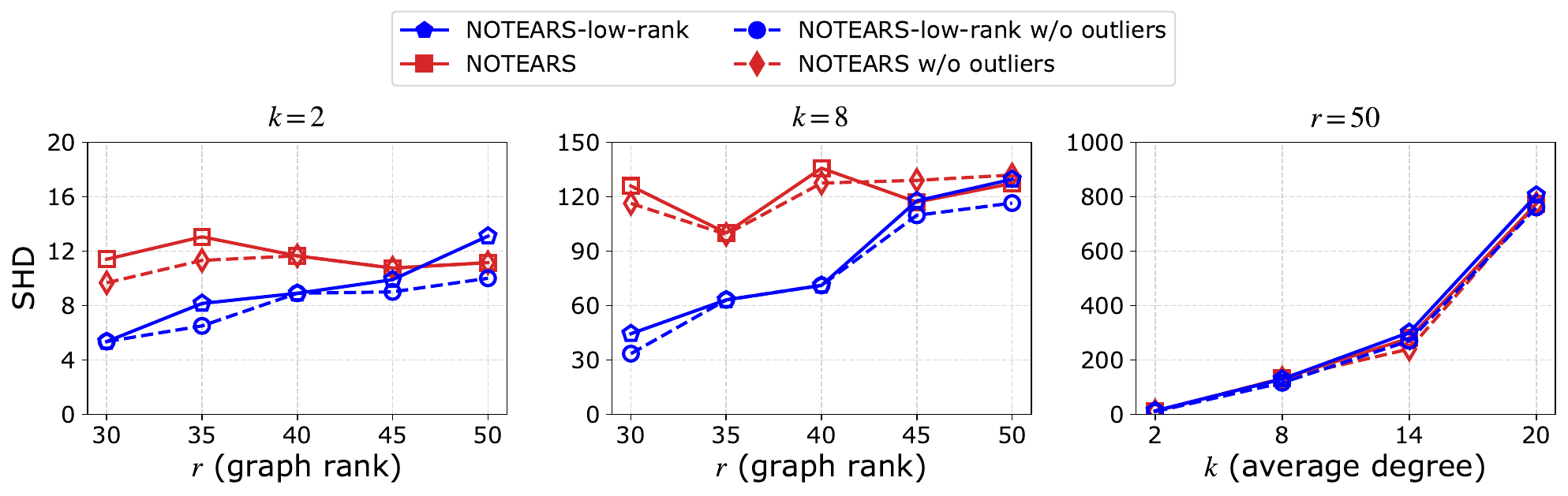}
 				\vspace{-1em}
 				\caption{Average SHDs on rank-specified graphs with higher ranks. }
 				\label{fig:other_sems}
 			\end{minipage}%
 		}%
 		\vspace{-.5em}
 	 	\hspace{1.5em}
 	\subfigure{
 			\begin{minipage}[t]{0.26\textwidth}
 				\centering	\includegraphics[width=\textwidth]{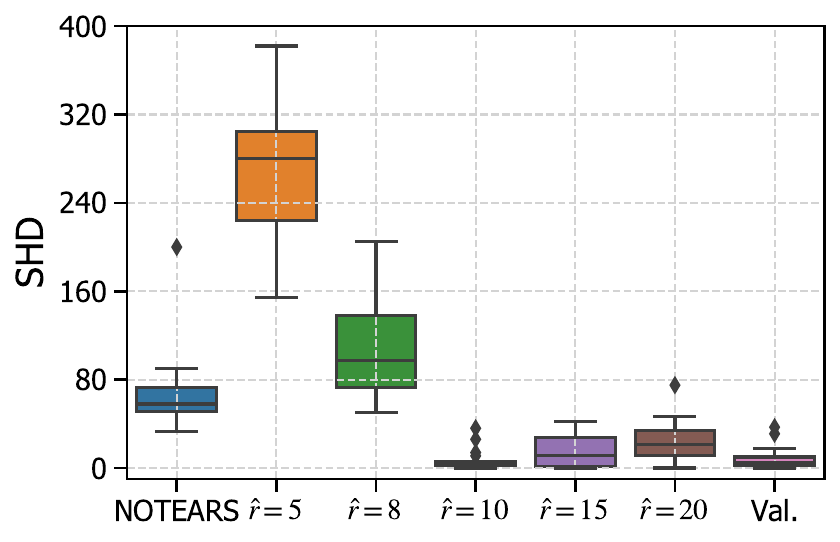}
 				\vspace{-1em}
 				\caption{Different rank parameters.}
 				\label{fig:sensitivity}
 			\end{minipage}%
 		}%
\vspace{-.5em}
\end{figure*}

{\bf Larger Graph} \quad Finally, we test the proposed approach for larger problems with $800$ nodes, average degree $8$, and rank $80$. We run GOLEM and its low rank version due to its computational efficiency for large problems. GOLEM achieves SHD $3698.9\pm 30.9$, compared with $284.2\pm380.0$ for GOLEM-low-rank. The high variance is due to a dataset where GOLEM-low-rank performs poorly. Excluding this dataset gives SHD being $158.2 \pm 140.2$.

\subsection{Sensitivity of Rank Parameters and Validation}
\label{sec:sentivity}

The previous section assumed that the true rank is known, serving as an initial check of the low rank modification.  In this experiment, we conduct an empirical analysis with different rank parameters $\hat r\in\{5,8,10,15,20\}$ for linear Gaussian data model on rank-specified graphs with $100$ nodes, degree $8$, and rank $10$. As reported in Figure~\ref{fig:sensitivity}, NOTEARS-low-rank performs the best when the rank parameter is identical to the true rank. Compared with  NOTEARS on the same datasets, the low rank version performs well across a range of rank parameters.

For our purpose, we further include the validation based approach that was described in Section~\ref{sec:learning}. We choose $2,000$ samples as training dataset and the rest as validation dataset. We assume only a range of possible ranks available, and here we simply consider that the rank parameters are also taken from $\{5,8,10,15,20\}$. We apply NOTEARS-low-rank with each of them and then evaluate each learned DAG using the validation dataset. The DAG with the best score would be selected as our estimate. We see that the validation based approach has almost the same performance as that of the true rank. Although this validation approach increases the total running time which depends on the number of candidate rank parameters, we believe it acceptable given the gain in accuracy and the fact that this strategy is commonly adopted for tuning other hyperparameters like the $\ell_1$ penalty weight.

\subsection{Linear Data Models with Scale-Free Graphs}
\label{sec:scale_free_exp}
We continue to consider scale-free graphs with $d=100$ nodes, average degrees $k\in\{6, 8\}$, and power $\gamma=2.5$. Such graphs generally have hubs, and if we know \emph{a priori} a rough estimate of $k$ and $\gamma$, we may use simulations (e.g., Figure~\ref{fig:sfer}) to infer the range of possible ranks with high confidence. In this experiment, we pick $\hat r\in\{15, 20, 25,30,35,40\}$ for NOTEARS-low-rank. For clarity of presentation, we only show the experimental results of some rank parameters in Figure~\ref{fig:sf}, along with the validation based approach. The low rank adaptations again results in an improved performance for causal structure learning. We also note that the validation based approach achieves a competitive result. Using Student’s t-test, the validation based approach is better than NOTEARS, with significance level $0.1$.

\setcounter{figure}{5}
	\begin{figure}[h]
		\subfigure{
			\begin{minipage}[t]{0.48\textwidth}
				\centering
	            \includegraphics[width=\textwidth]{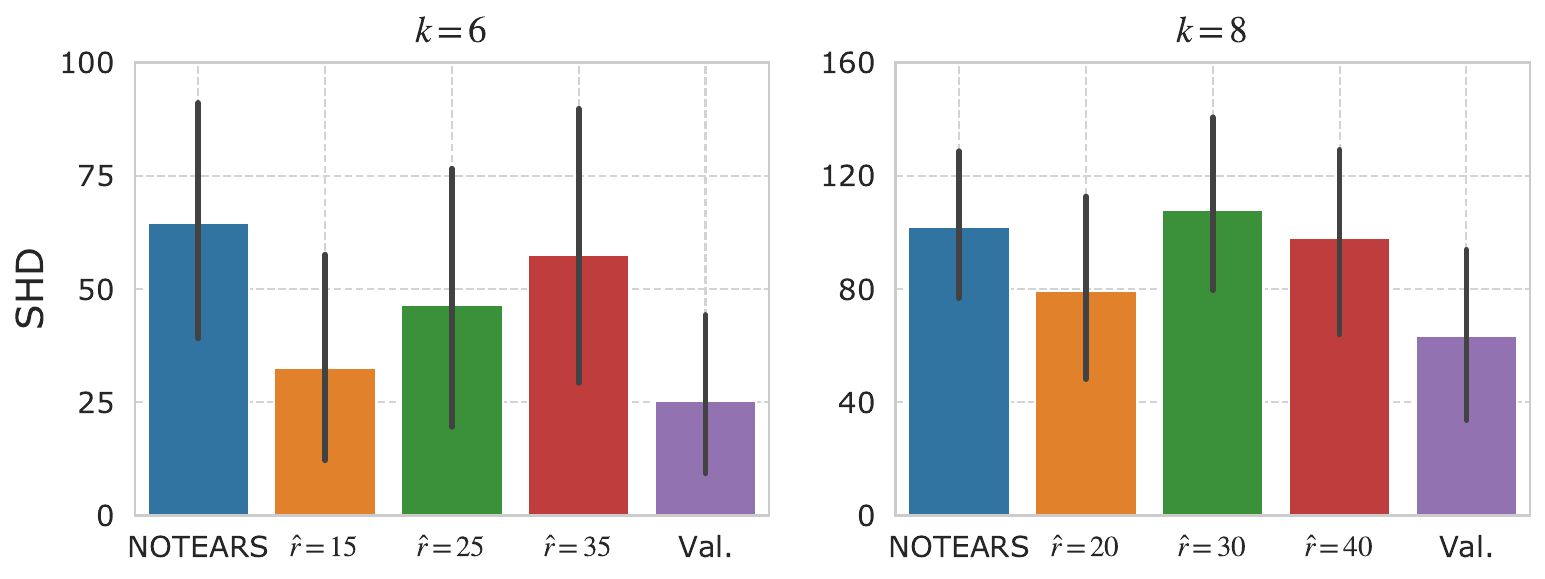}
			\end{minipage}%
		}%
		\vspace{-0.5em}
		\caption{Scale-free graphs.} 
		\label{fig:sf}
		\vspace{-0.5em}
	\end{figure}

	\subsection{Non-Linear SCMs} 
	\label{nonlinearSEMs}
	
\begin{figure}[h]
		\centering
		\includegraphics[width=0.48\textwidth]{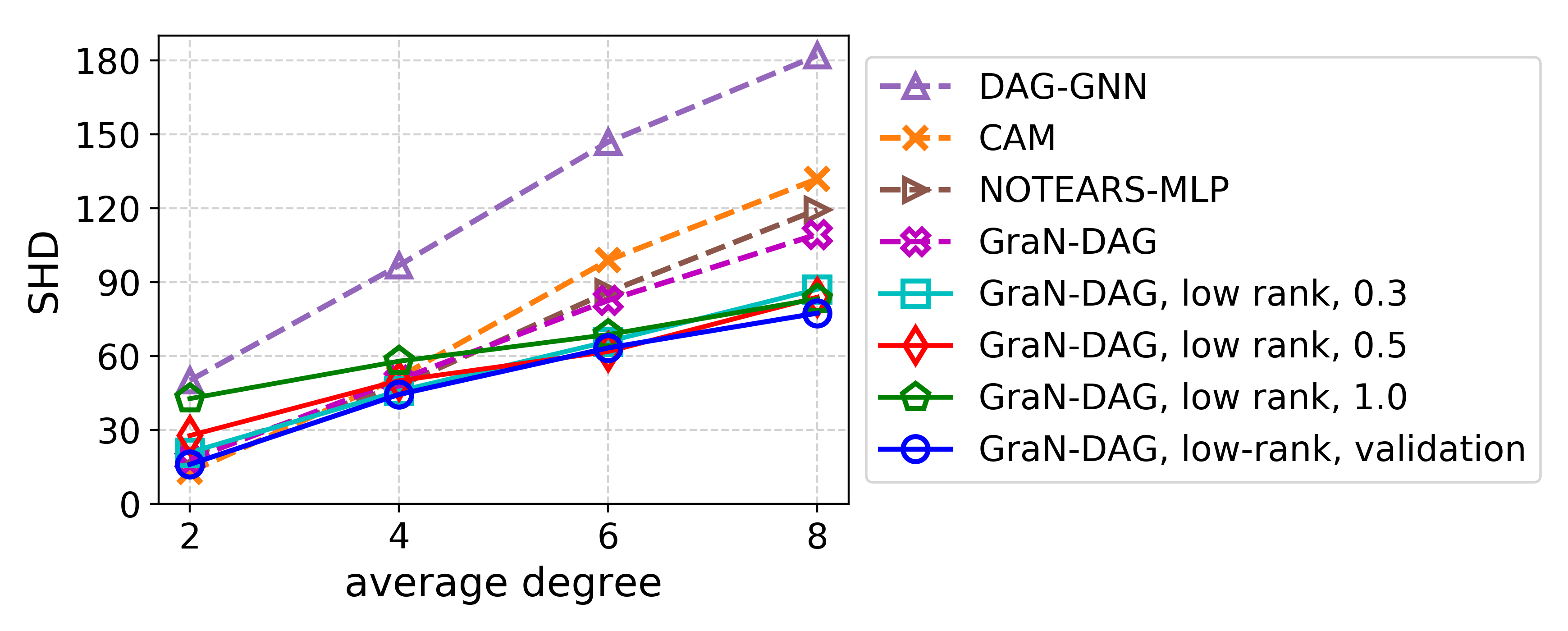}
		\vspace{-0.5em}
		\caption{Non-linear SCMs.}
		\label{fig:nonlinear}
		\vspace{-.5em}
        \end{figure}

	For non-linear data models, we pick rank-specified graphs with $50$ nodes, rank $5$, and average degree $k\in\{2,4,6,8\}$. To our knowledge, the selected benchmark methods CAM, NOTEARS-MLP, and GraN-DAG are  state-of-the-art methods on this data model which is known to be identifiable \cite{Peters2014causal}.  As a demonstration of the low rank assumption, we apply the nuclear norm approach to GraN-DAG and choose from $\{0.3, 0.5, 1\}$ as penalty weights. For the validation approach, we use the same splitting ratio as in Section~\ref{sec:sentivity} and consider more penalty weights from $\{0.1, 0.3, 0.5, 0.8, 1, 2, 5\}$. The learned graph that achieves the best score on the validation dataset is chosen as final estimate. Figure~\ref{fig:nonlinear} (and Appendix~\ref{app:detaile_nonlinear} with more detailed results) shows that adding a nuclear norm can  improve the performance of GraN-DAG across a large range of weights when the graph is relatively dense. For degree $8$, the low rank version with validation achieves average SHD  $77.4$, while the SHDs of CAM, NOTEARS-MLP, and original GraN-DAG are $131.9$, $119.4$, and $109.4$,  respectively.
	
We further check how the low rank method behaves when the underlying graph is not low rank. We consider regular  ER graphs with $50$ nodes and the same data models. With a large penalty weight, the learned DAGs tend to have high SHDs in this case. The original GraN-DAG achieves SHD $3.4\pm 2.1$ and $44.2\pm 12.4$ for average degrees $2$ and $8$, respectively, while the validation based approach, including original GraN-DAG, has similar SHDs $2.2\pm 1.2$ and $48.2\pm 11.2$, respectively. Indeed, the final estimates are mostly obtained by original GraN-DAG or the low rank version with small penalty weights.

	\subsection{Real Network}
	\label{sec:real}
\begin{figure}[t]
		\begin{center}
			\includegraphics[width=0.27\textwidth]{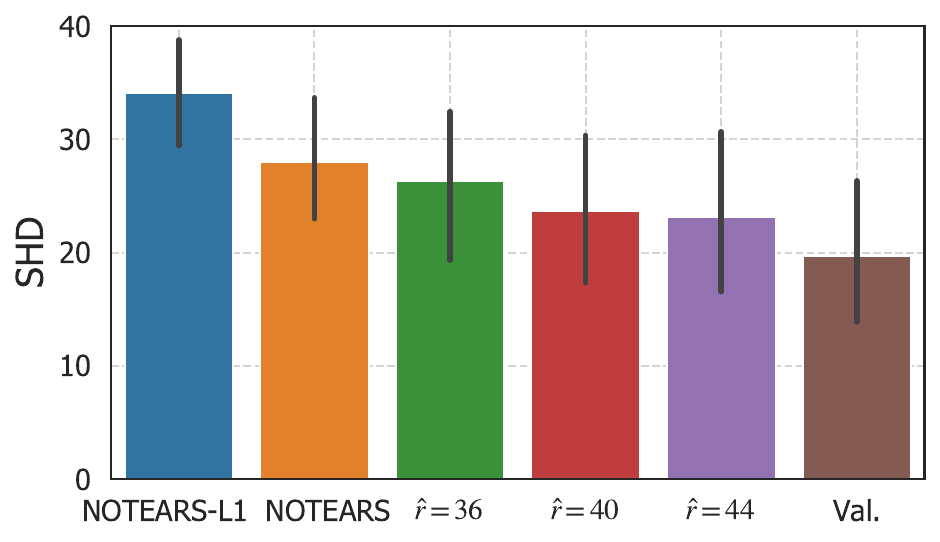}
		\end{center}
		\vspace{-0.5em}
		\caption{Real network: {\it arth150}.}
		\label{fig:real}
	\end{figure}

	We apply the proposed method to the \emph{arth150} gene network available at the Bayesian network repository,\footnote{{https://www.bnlearn.com/bnrepository}} which is a DAG containing $107$ genes and $150$ edges. The maximum rank is $40$ and maximum in-degree is $6$. Since the real dataset  has only $22$ samples, we instead use simulated data from linear data model with standard Gaussian noises. We pick $\hat{r}$ from $\{36, 40, 44\}$ and  also use validation to select the rank parameter. We  apply NOTEARS-L1 where the penalty weight is chosen from $\{0.05, 0.1, 0.2\}$, and similarly treat this method favorably by picking the lowest SHD for each dataset. The empirical results are shown in Figure~\ref{fig:real}. Using Student's t-test, we find that with significance level $0.1$, the results obtained with $\hat{r}=40$ and the validation approach are better than NOTEARS. Notice that  the true rank is not very low and the maximum in-degree is small. Even so, the low rank adaptation remains useful.
	
\begin{figure}[t]
		\begin{center}
			\includegraphics[width=0.25\textwidth]{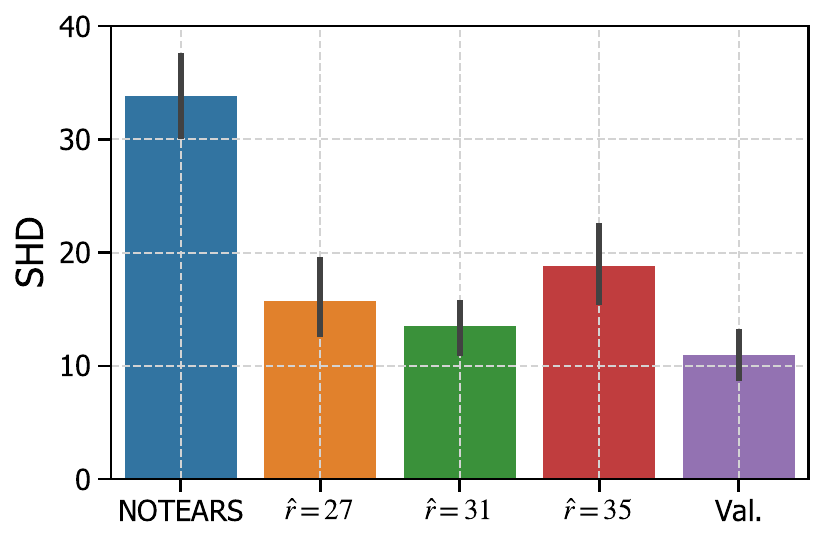}
		\end{center}
		\vspace{-0.5em}
		\caption{Real network: {\it pathfinder}.}
		\label{fig:pathfinder}
		\vspace{-.5em}
	\end{figure}

	We next consider another benchmark network \emph{pathfinder}, also from the Bayesian network repository. The  network has $109$ nodes and $195$ edges, and the average degree is $3.57$. The maximum rank of this graph is $31$, and the lower bound on the minimum rank computed using our proposed method is $26$. We adopt a similar experimental setup and pick $\hat{r}$ from $\{27, 31, 35\}$. As reported in Figure~\ref{fig:pathfinder}, the low rank version has a much lower SHD than original NOTEARS.

	\section{Concluding Remarks}\label{sec:conclusion}
	{Causal DAG models have been  applied in many fields of machine learning and data science~\cite{Bernhard2021Toward, Bernhard2022Causality}, such as computer vision~\cite{Kocaoglu2017causalgan} and algorithmic fairness~\cite{NIPS2017_a486cd07}, to improve  explainability and interpretability.} This paper studies the problem of learning DAGs from observational data and makes a first step towards incorporating low-rankness into causal DAG learning, with which we aim to deal with relatively large and dense causal DAGs. Empirically, we show that the low rank adaptations perform noticeably better than existing algorithms when the low rank condition is  satisfied, and also deliver competitive performance when the rank is not restricted to be low. Our theoretical results provide a better understanding of low rank graphs in terms of graphical features and hence of the low rank assumption itself. 

\appendix

\subsection{Proofs}\label{app:app:proof}
		
	For ease of presentation, we occasionally use index $i$ to represent variable $X_i$ in the following sections.
	
	\subsubsection{Proof of Theorem \ref{thm:lower}}
		Let ${\cal G}=(\mathbf{V}, \mathbf{E})$. Consider an equivalence relation, denoted by $\sim$, among vertices in $\mathbf{V}$ defined as follows: for any $X_i, X_j\in \mathbf{V}$, $X_i\sim X_j$ if and only if $l(X_i)=l(X_j)$ and $X_i$ and $X_j$ are connected. Here, connected means that there is a path between $X_i$ and $X_j$. Below we use $C(X_i)$ to denote the equivalence class containing $X_i$. Next, we define a weak ordering $\pi$ on $\mathbf{V}/\sim$, i.e., the equivalence classes induced by $\sim$, by letting $C(X_i) \succeq_\pi C(X_j)$ if and only if $l(X_i)\geq l(X_j)$. Then, we extend $\succeq_\pi$ to a total ordering $\rho$ on $\mathbf{V}/\sim$. The ordering $\rho$ also induces a weak ordering (denoted by $\Bar{\rho}$) on $\mathbf{V}$: $X_i\succeq_{\Bar{\rho}} X_j$ if and only if $C(X_i) \succ_\rho C(X_j)$. Finally, we extend $\Bar{\rho}$ to a total ordering $\gamma$ on $\mathbf{V}$. It can be verified that $\gamma$ is a topological ordering of ${\cal G}$, that is, if we relabel the vertices according to $\gamma$, then $X_i\in \mathrm{pa}(X_j, {\cal G})$ if and only if $i > j$ and $X_i$ and $X_j$ are adjacent, and the adjacency matrix of ${\cal G}$ becomes lower triangular.
		
		Assume that the vertices of ${\cal G}$ are relabeled according to $\gamma$ and we will consider the binary adjacency matrix $A$ of the resulting graph throughout the rest of this proof. Note that relabelling is equivalent to applying a permutation onto the adjacency matrix, which does not change the rank. Let $\mathbf{V}_{0}=\{1,2,\ldots,k_1-1\}$ for some $k_1\geq 2$. Then the $k_1$-th row of $A$, denoted by $A(k_1,\cdot)$, is the first non-zero row vector of $A$. Letting $S=\{A(k_1,\cdot)\}$, then $S$ contains a subset of linearly independent vector(s) of the first $k_1$ rows of $A$. Suppose that we have visited the first $m$ rows of $A$ and $S=\{A(k_1,\cdot),A(k_2,\cdot),\ldots,A(k_t,\cdot)\}$ contains a subset of linearly independent vector(s) of the first $m$ rows of $A$, where $k_1\leq m<d$. If $X_{m+1}\nsim X_{k_t}$, then we add $A(m+1,\cdot)$ to $S$; otherwise, we keep $S$ unchanged. We claim that the vectors in $S$ are still linearly independent after the above step. Clearly, if we do not add any new vector, then $S$ contains only linearly independent vectors. To show the other case, note that if $l(X_{m+1})>l(X_{k_t})\geq\cdots\geq l(X_{k_1})$, then  there is an index $i\in \mathbf{V}_{l(X_{m+1})-1}$ such that $A(m+1, i)\neq0$, by the definition of height. Since $l(X_{m+1})>l(X_{k_t})$, we have $l(X_{k_t})\leq l(X_{m+1})-1$ and thus $A(k_j, i)=0$ for all $j=1,2,\ldots,t$. Therefore, $A(m+1,\cdot)$ cannot be linearly represented by $\{A(k_j, \cdot); j=1,2,\ldots, t\}$ and the vectors in $S$ are linearly independent. On the other hand, if $l(X_{m+1})=l(X_{k_t})$, then the definition of the equivalence relation $\sim$ implies that $X_{m+1}$ and $X_{k_t}$ are disconnected, which means that $X_{m+1}$ and $X_{k_t}$ do not share a common child in $\mathbf{V}_{l(X_{m+1})-1}$. Consequently, there is an index $i\in\mathbf{V}_{l(X_{m+1})-1}$ such that $A(m+1, i)\neq0$ but $A(k_t, i)=0$. Similarly, we can show that $A(k_j, i)=0$ for all $j=1,2,\ldots,t$. Thus, the vectors in $S$ are still linearly independent.
		
		After visiting all the rows in $A$, the number of vectors in $S$ is equal to $\sum\nolimits_{s=1}^{l({\cal G})} |\mathcal{C}(\mathcal{G}_{s,s-1})|$ based on the definition of $\sim$.  The second inequality can be shown by noting that $\mathcal{C}(\mathcal{G}_{s,s-1})$ has at least one elements. The proof is complete.
	
	\subsubsection{Proof of Theorem \ref{thm:matching}}
		Denote the directed graph by ${\cal G}=(\mathbf{V}, \mathbf{E})$. \cite[Theorem~1]{edmonds1967systems} showed that $\max\{\operatorname{rank}(W); W\in\mathcal{W}_A\}$ is equal to the maximum number of nonzero entries of $A$, no two of which lie in a common row or column. Therefore, it suffices to show that the latter quantity  is equal to the size of the minimum head-tail vertex cover. Let $\mathbf{V}^{'} = \mathbf{V}'_0\cup\mathbf{V}'_1$, where  $\mathbf{V}'_0=\mathbf{V}\times\{0\}=\{(X_i, 0); X_i\in\mathbf{V}\}$ and $\mathbf{V}'_1=\mathbf{V}\times\{1\}=\{(X_i, 1); X_i\in\mathbf{V}\}$. Now define a bipartite graph ${\cal B}=(\mathbf{V}^{'}, \mathbf{E}^{'})$ where $\mathbf{E}^{'}=\{(X_i, 0)\to (X_j,1); (X_i,X_j)\in \mathbf{E}\}$. Denote by $\cal M$ a set of nonzero entries of $A$ so that no two entries lie in the same row or column. Notice that $\cal M$ can  be viewed as an edge set and no two edges in $\cal M$ share a common endpoint. Thus, $\cal M$ is a matching of ${\cal B}$. Conversely, it can be shown by similar arguments that any matching of ${\cal B}$ corresponds to a set of nonzero entries of $A$, no two of which lie in a common row or column. Therefore, $\max\{\operatorname{rank}(W), W\in\mathcal{W}_A\}$ equals the size of the maximum matching of ${\cal B}$, and further the size of the minimum vertex cover of ${\cal B}$ according to K{\"o}nig's theorem. Note that any vertex cover of ${\cal B}$ can be equivalently transformed to a head-tail vertex cover of $\cal G$, by letting $\mathbf{H}$ and $\mathbf T$ be the subsets of the vertex cover containing all variables in $\mathbf{V}'_0$ and of the vertex cover containing all variables in $\mathbf{V}'_1$, respectively. Thus, $\max\{\operatorname{rank}(W), W\in\mathcal{W}_A\}$ is equal to the size of the minimum head-tail vertex cover.
	
	\subsubsection{Proof of Theorem \ref{thm:upper-loose}}
		We start with the first inequality in Proof of Theorem \ref{thm:upper-loose}. Let $h_1,\cdots,h_p$ denote the heights where $|\mathbf{V}_s| <  |\mathrm{ch}(\mathbf{V}_s)|$, and $t_1,\cdots,t_q$  the heights where $|\mathbf{V}_s| > |\mathrm{ch}(\mathbf{V}_s)|$. Let $\mathbf{H}=\cup_{i=1}^p \mathbf{V}_{h_i}$ and $\mathbf{T}=\cup_{i=1}^q \mathbf{V}_{t_i}$. It is straightforward to see that $(\mathbf{H},\mathbf{T})$ is a head-tail vertex cover. Thus, Proof of Theorem \ref{thm:upper-loose} holds  according to Theorem \ref{thm:matching}. The second inequality can be shown similarly and its proof is omitted. For the third inequality, let $m=\mathrm{arg max}\{|\mathbf{V}_s|: 0\leq s\leq {l({\cal G})}\}$, and define $\mathbf{H}=\cup_{i>m}\mathbf{V}_{i}$ and $\mathbf{T}=\cup_{i<m}\mathbf{V}_{i}$. Then $(\mathbf{H},\mathbf{T})$ is also a head-tail vertex cover and the third inequality follows from Theorem~\ref{thm:matching}, too.
	
	\subsubsection{Proof of Theorem \ref{thm:upper-undirected}}
		Notice that Theorem~\ref{thm:matching} holds for all directed graphs. This theorem then follows by treating the skeleton and the moral graph  as directed graphs with loops, i.e., an undirected edge $X_i - X_j$ is treated as two directed edges $X_i\to X_j$ and $X_j\to X_i$.
	
		\begin{table*}[t!]
		\centering
		\caption{Detailed results for linear Gaussian data model with equal noise variances. Time is in minutes, unless otherwise stated.}
		\resizebox{\linewidth}{!}{
		{
			\begin{tabular}{lc|cccc|cccc}
				\toprule
				&~& \multicolumn{4}{c|}{\bf 100 nodes, rank 10} & \multicolumn{4}{c}{\bf 300 nodes, rank 30} \\
				\midrule 
				& Degree & { 2} & { 4} & { 6} & { 8} & { 2} & { 4} & { 6} & { 8}
				\\ \midrule
				\multirowcell{4}{NOTEARS-\\low-rank}
				&TPR     &    0.99\,$\pm$\,0.02  & 0.99\,$\pm$\,0.03   & 0.99\,$\pm$\,0.01  &  0.99\,$\pm$\,0.01 
				    &  0.99\,$\pm$\,0.01 & 0.99\,$\pm$\,0.01 &0.99\,$\pm$\,0.01& 0.97\,$\pm$\,0.05\\
				&FDR    &   0.001\,$\pm$\,0.003  & 0.01\,$\pm$\,0.02  & 0.02\,$\pm$\,0.04   &  0.02\,$\pm$\,0.02 \
				    &  0.03\,$\pm$\,0.03 & 0.02\,$\pm$\,0.02 & 0.04\,$\pm$\,0.02& 0.06\,$\pm$\,0.08 \\
				&SHD    &    1.3\,$\pm$\,3.5 &4.3\,$\pm$\,8.7 &\,~9.8\,$\pm$\,13.9 &14.5\,$\pm$\,18.2
				     & 6.4\,$\pm$\,8.4 &14.6\,$\pm$\,15.1 &32.6\,$\pm$\,29.7 &\,~88.0\,$\pm$\,137.0 \\
				&Time     &   1.9\,$\pm$\,0.2 & 5.7\,$\pm$\,2.7   & 5.8\,$\pm$\,2.6  &  7.5\,$\pm$\,2.7 &  57.6\,$\pm$\,43.2 & 76.5\,$\pm$\,27.8 & 158.7\,$\pm$\,94.7~~& 262.9\,$\pm$\,144.4\\
				\midrule
					\multirowcell{4}{GOLEM-\\low-rank}&TPR     &     0.99\,$\pm$\,0.01 & 0.99\,$\pm$\,0.01 &0.99\,$\pm$\,0.01& 0.97\,$\pm$\,0.05 &    0.998\,$\pm$\,0.004 & 0.995\,$\pm$\,0.008 & 0.95\,$\pm$\,0.10 & 0.93\,$\pm$\,0.11\\
				&FDR    &   0.03\,$\pm$\,0.03 & 0.02\,$\pm$\,0.02 & 0.04\,$\pm$\,0.02& 0.06\,$\pm$\,0.08 &   0.001\,$\pm$\,0.002 & 0.01\,$\pm$\,0.01 & 0.05\,$\pm$\,0.03 & 0.11\,$\pm$\,0.03\\
				&SHD    &    0.4\,$\pm$\,0.9 & 2.3\,$\pm$\,3.2 & 5.5\,$\pm$\,5.7 & 21.6\,$\pm$\,24.5 &   1.6\,$\pm$\,3.9 & 4.4\,$\pm$\,6.7 & \,~74.9\,$\pm$\,171.3 & 105.1\,$\pm$\,179.6\\
				&Time    &3.9\,$\pm$\,0.2&4.0\,$\pm$\,0.3 & 4.0\,$\pm$\,0.1 & 4.6\,$\pm$\,1.0&26.9\,$\pm$\,0.2\,~ & 27.8\,$\pm$\,0.2~~& 36.7\,$\pm$\,2.4\,~ &  35.1\,$\pm$\,7.3\,~  \\
				\midrule
				\multirowcell{4}{NOTEARS}&TPR&    0.90\,$\pm$\,0.06 & 0.90\,$\pm$\,0.04 &   0.87\,$\pm$\,0.05 & 0.89\,$\pm$\,0.03  &   0.94\,$\pm$\,0.01 & 0.93\,$\pm$\,0.02 &  0.93\,$\pm$\,0.02 & 0.91\,$\pm$\,0.02 \\
				&FDR   &    0.07\,$\pm$\,0.03  & 0.06\,$\pm$\,0.04  &  0.06\,$\pm$\,0.02   &  0.04\,$\pm$\,0.02&  0.07\,$\pm$\,0.02& 0.06\,$\pm$\,0.03 & 0.08\,$\pm$\,0.04& 0.09\,$\pm$\,0.06 \\
				&SHD        &    9.8\,$\pm$\,7.7 &26.2\,$\pm$\,12.5 &48.6\,$\pm$\,17.9 &58.7\,$\pm$\,15.8 &  26.6\,$\pm$\,10.4 &72.2\,$\pm$\,19.1 &133.8\,$\pm$\,43.3\,~~&186.5\,$\pm$\,85.0\,~ \\
				&Time &    3.6\,$\pm$\,1.6 & 6.6\,$\pm$\,1.6   & 9.6\,$\pm$\,1.5   &  7.3\,$\pm$\,1.5
				&  23.9\,$\pm$\,5.8\,~ & 42.0\,$\pm$\,10.5 & 74.2\,$\pm$\,37.9 & 104.2\,$\pm$\,21.7\,~\,\\
				\midrule
					\multirowcell{4}{GOLEM}&TPR     &  0.997\,$\pm$\,0.008& 0.99\,$\pm$\,0.01 & 0.99\,$\pm$\,0.02 & 0.95\,$\pm$\,0.06 &   0.98\,$\pm$\,0.02 & 0.86\,$\pm$\,0.04 & 0.71\,$\pm$\,0.06 & 0.55\,$\pm$\,0.05\\
				&FDR    &   0.001\,$\pm$\,0.003 & 0.005\,$\pm$\,0.007& 0.01\,$\pm$\,0.01 & 0.01\,$\pm$\,0.01 &   0.001\,$\pm$\,0.002 & 0.01\,$\pm$\,0.01 & 0.05\,$\pm$\,0.03 & 0.11\,$\pm$\,0.03\\
				&SHD    &    2.9\,$\pm$\,6.2 & 38.5\,$\pm$\,35.0 & 114.3\,$\pm$\,43.8\,~ & 185.5\,$\pm$\,85.6\,~  &    7.9\,$\pm$\,5.6 & 87.3\,$\pm$\,30.1 & 293.2\,$\pm$\,84.3\,~ & 628.5\,$\pm$\,59.9\,~ \\
				&Time     & 3.6\,$\pm$\,0.1& 3.7\,$\pm$\,0.1& 3.8\,$\pm$\,0.1& 3.7\,$\pm$\,0.2 & 42.8\,$\pm$\,1.8\,~ & 42.8\,$\pm$\,1.8\,~~& 28.1\,$\pm$\,0.7\,~   & 28.0\,$\pm$\,0.6\,~ \\
				\midrule
				\multirowcell{4}{fast GES}& TPR &0.31\,$\pm$\,0.04 &0.13\,$\pm$\,0.02 &0.08\,$\pm$\,0.01 &0.06\,$\pm$\,0.01 & 0.32\,$\pm$\,0.04 &0.13\,$\pm$\,0.01 &0.07\,$\pm$\,0.01 &0.05\,$\pm$\,0.01 \\
& FDR &0.75\,$\pm$\,0.03 &0.84\,$\pm$\,0.02 &0.87\,$\pm$\,0.02 &0.88\,$\pm$\,0.01 &  0.77\,$\pm$\,0.02 &0.84\,$\pm$\,0.01 &0.87\,$\pm$\,0.01 &0.89\,$\pm$\,0.01 \\
& SHD &158.1\,$\pm$\,17.8\,~ &304.8\,$\pm$\,25.4\,~ &426.9\,$\pm$\,23.6\,~ &534.8\,$\pm$\,26.3\,~ & 532.5\,$\pm$\,38.2\,~ &917.6\,$\pm$\,34.9\,~~&1281.6\,$\pm$\,48.5~~~~&1613.8\,$\pm$\,54.9~~~~\\
				&Time & \multicolumn{4}{c|}{$<$ 10 seconds}  & \multicolumn{4}{c}{$<$ 30 seconds} \\
				\midrule
					\multirowcell{4}{PC} & TPR &0.65\,$\pm$\,0.07 &0.28\,$\pm$\,0.09 &0.15\,$\pm$\,0.04 &0.09\,$\pm$\,0.02 & 0.69\,$\pm$\,0.06 &0.30\,$\pm$\,0.11 &0.17\,$\pm$\,0.03 &0.10\,$\pm$\,0.02  \\
& FDR &0.31\,$\pm$\,0.06 &0.50\,$\pm$\,0.16 &0.71\,$\pm$\,0.08 &0.81\,$\pm$\,0.05 & 0.38\,$\pm$\,0.05 &0.45\,$\pm$\,0.16 &0.66\,$\pm$\,0.06 &0.77\,$\pm$\,0.04\\
& SHD &52.1\,$\pm$\,11.4 &187.2\,$\pm$\,40.2\,~ &346.7\,$\pm$\,40.0\,~ &491.1\,$\pm$\,35.7\,~ & 181.4\,$\pm$\,31.5 \,~&531.7\,$\pm$\,60.0\,~~&995.8\,$\pm$\,84.8\,~~&1420.5\,$\pm$\,89.6~~~~\\
				&Time     &  \multicolumn{4}{c|}{$<$ 30 seconds} & \multicolumn{4}{c}{$<$ 90 seconds}\\
				\midrule
					\multirowcell{4}{MMHC}& TPR &0.78\,$\pm$\,0.08 &0.39\,$\pm$\,0.08 &0.20\,$\pm$\,0.02 &0.11\,$\pm$\,0.01 & \multicolumn{4}{c}{N/A}\\
& FDR &0.17\,$\pm$\,0.05 &0.30\,$\pm$\,0.09 &0.55\,$\pm$\,0.04 &0.71\,$\pm$\,0.02 & \multicolumn{4}{c}{N/A}  \\
& SHD &37.6\,$\pm$\,10.0 &152.6\,$\pm$\,32.1\,~~&308.3\,$\pm$\,27.7\,~ &447.0\,$\pm$\,28.6~ & \multicolumn{4}{c}{N/A}\\
				&Time     & 1.00\,$\pm$\,0.47 & 10.46\,$\pm$\,23.49 &\multicolumn{2}{c|}{around 20 hours} & \multicolumn{4}{c}{$>$ 24 hours} \\
				\midrule
					\multirowcell{4}{Method \\ from \cite{Ghoshal2018learning}}& TPR &0.99\,$\pm$\,0.02 &0.83\,$\pm$\,0.14 &0.54\,$\pm$\,0.15 &0.28\,$\pm$\,0.09 & 0.96\,$\pm$\,0.04 &0.52\,$\pm$\,0.13 &0.13\,$\pm$\,0.05 &0.04\,$\pm$\,0.02 \\
& FDR &0.001\,$\pm$\,0.03\,~ &0.03\,$\pm$\,0.06 &0.10\,$\pm$\,0.10 &0.15\,$\pm$\,0.10 & 0.01\,$\pm$\,0.02 &0.09\,$\pm$\,0.06 &0.31\,$\pm$\,0.18 &0.62\,$\pm$\,0.14\\
& SHD &1.3\,$\pm$\,2.4 &40.3\,$\pm$\,40.2&155.7\,$\pm$\,58.8\,~~&307.3\,$\pm$\,52.7\,~ & 15.4\,$\pm$\,16.6 &317.9\,$\pm$\,89.2\,~ &832.5\,$\pm$\,77.1 \,~&1235.4\,$\pm$\,64.3~~~ \\
				&Time     &  \multicolumn{4}{c|}{$<$ 5 seconds} & \multicolumn{4}{c}{$<$ 60 seconds}\\
				\bottomrule
				\vspace{0.5em}
			\end{tabular}
		}
		}
		\label{tab:detail}
	\end{table*}
	\begin{table*}[t!]
		\centering
		\caption{Detailed SHDs for Experiment~4 with non-linear SEMs.}
		\setlength{\tabcolsep}{4mm}{
		\scalebox{0.95}{
		{
			\begin{tabular}{lccccccc}
				\toprule
				Degree & {2} & {4} & {6} & {8}
				\\
				\midrule
				DAG-GNN    &   47.4\,$\pm$\,8.9&96.5\,$\pm$\,11.6 & 146.5\,$\pm$\,10.6~\,&177.2\,$\pm$\,15.9\\
				CAM       &    {\bf 13.8\,$\pm$\,4.4} &50.0\,$\pm$\,17.3 & 97.2\,$\pm$\,19.2&130.7\,$\pm$\,27.7 \\
				NOTEARS-MLP  & 24.9\,$\pm$\,5.7 & {\bf 59.2\,$\pm$\,18.9} &
				92.7\,$\pm$\,22.4&
				128.1\,$\pm$\,26.6 \\
				GraN-DAG  & \,~17.9\,$\pm$\,17.0&50.3\,$\pm$\,51.7& {\bf 82.6\,$\pm$\,75.8}& \,~{\bf 109.4\,$\pm$\,102.4}
				\\
				\midrule
				GraN-DAG, low rank, $0.3$  &20.9\,$\pm$\,23.2&45.8\,$\pm$\,47.7&65.9\,$\pm$\,59.1&87.1\,$\pm$\,79.9\\
				GraN-DAG, low rank, $0.5$    & 27.7\,$\pm$\,40.8&50.0\,$\pm$\,53.4&{\bf 61.8\,$\pm$\,66.7}&83.9\,$\pm$\,85.2\\
				GraN-DAG, low rank, $1.0$    & 42.7\,$\pm$\,58.8&57.9\,$\pm$\,67.6&68.7\,$\pm$\,76.2&83.2\,$\pm$\,76.9\\
				GraN-DAG, low rank, validation~~~~    & {\bf 16.0\,$\pm$\,4.5}\,~ &
				{\bf 44.4\,$\pm$\,21.0}&
				63.3\,$\pm$\,24.7&
				{\bf 77.4\,$\pm$\,28.8}\\
				\bottomrule
		\end{tabular}}}}
		\label{tab:detail33}
	\end{table*}

\subsection{Detailed Empirical Results for Experiment~1 with Linear Gaussian SEMs}
\label{app:detailed_result}
Table~\ref{tab:detail} reports detailed results including true positive rates (TPRs), false discovery rates (FDRs), structural Hamming distances (SHDs), and running time on rank-specified graphs with linear data models where the noises follow standard Gaussian distribution. Here the true rank is assumed to be known and is used as the rank parameter in NOTEARS-low-rank and GOLEM-low-rank. For both GOLEM and GOLEM-low-rank, we set $30,000$ and $100,000$ optimization steps in the stochastic first-order method for $100$- and $300$-node graphs, respectively. We also include (fast) GES, MMHC, and PC. However, PC is too slow since some nodes may have a high in-degree (i.e., hubs) in large, dense, and low rank graphs. For the same reason, the skeleton may not be correctly estimated by MMHC, which has a similar performance to that of  GES. Therefore, we only include the results of GES for comparison. We treat GES favorably by regarding  undirected edges as true positives if the true graph has a directed edge in place of the undirected ones. 

\subsection{Detailed  Results for Experiment~4 with Non-Linear SEMs}
\label{app:detaile_nonlinear}
Table~\ref{tab:detail33} reports the detailed SHDs for each method in Section~\ref{nonlinearSEMs}. We also mark in bold the best results from  methods with or without low rank modifications.

	
\bibliographystyle{IEEEtran}	
\bibliography{szhu}

\end{document}